\documentclass[journal]{IEEEtran}

\usepackage[space, compress, sort]{cite}
\usepackage{amsmath}
\usepackage{amsfonts}

\usepackage{amsthm}
\usepackage[ruled,lined]{algorithm2e}
\usepackage{mathtools}
\usepackage{tabularx}
\usepackage{tabularx}
\usepackage{graphicx}
\usepackage{enumerate}
\usepackage{float}
\usepackage{url}
\usepackage{verbatim}
\usepackage{booktabs} 
\usepackage{multirow}
\usepackage[linkcolor=black,citecolor=black,urlcolor=black,colorlinks=true]{hyperref}
\usepackage{bm}
\usepackage[normalem]{ulem}
\usepackage{soul}
\usepackage{xcolor}
\usepackage{etoolbox}
\usepackage{setspace}
\usepackage{amsthm}
\usepackage{amssymb,stmaryrd}
\usepackage{bm}
\usepackage{caption}
\usepackage{subcaption}
\usepackage{makecell}
\usepackage[mathscr]{euscript}

\graphicspath{{figures/}}
\IEEEoverridecommandlockouts

\author{  Zhichao Han$^{*1,2}$, Yuwei Wu$^{*3}$, Tong Li$^{4}$, Lu Zhang$^{4}$, Liuao Pei$^{1,2}$,\\
	 Long Xu$^{1,2}$, Chengyang Li$^{2}$, Changjia Ma$^{1,2}$, Chao Xu$^{1,2}$, Shaojie Shen$^{4}$, and Fei Gao$^{1,2}$
	\thanks{*  co-first authors.}
	\thanks{$^{1}$State Key Laboratory of Industrial Control Technology, Zhejiang University, Hangzhou 310027, China. (\textit{Corresponding author: Fei Gao})
	}
	\thanks{$^{2}$Huzhou Institute of Zhejiang University, Huzhou 313000, China.
	}
	\thanks{$^{3}$Department of Electrical and Systems Engineering, University of Pennsylvania, Philadelphia, PA 19104 USA.}
	\thanks{$^{4}$Department of Electronic and Computer Engineering, Hong Kong University of Science and Technology, Hong Kong.}
	\thanks{Yuwei Wu contributes to this work during internship at Zhejiang University.}
	
	\thanks{E-mail: {\tt\small \{ zhichaohan, fgaoaa\}@zju.edu.cn}}
}

\title{An Efficient Spatial-Temporal Trajectory Planner for Autonomous Vehicles in Unstructured Environments}

\begin{document}
    \maketitle

\begin{abstract}
As a core part of autonomous driving systems, motion planning has received extensive attention from academia and industry. 
However, real-time trajectory planning capable of spatial-temporal joint optimization is challenged by nonholonomic dynamics, particularly in the presence of unstructured environments and dynamic obstacles.
To bridge the gap, we propose a real-time trajectory optimization method that can generate a high-quality whole-body trajectory under arbitrary environmental constraints. By leveraging the differential flatness property of car-like robots, we simplify the trajectory representation and analytically formulate the planning problem while maintaining the feasibility of the nonholonomic dynamics. Moreover, we achieve efficient obstacle avoidance with a safe driving corridor for unmodelled obstacles and signed distance approximations for dynamic moving objects.
We present comprehensive benchmarks  with State-of-the-Art methods, demonstrating the  significance of the proposed method in terms of efficiency and trajectory quality. Real-world experiments verify the practicality of our algorithm.
We will release our codes for the research community.\footnote{\url{https://github.com/ZJU-FAST-Lab/Dftpav}}
\end{abstract}


	\section{Introduction}
\label{sec:introduction}
\begin{figure}	
	\centering
	\begin{subfigure}[t]{1.0\columnwidth}
		\centering
		\includegraphics[width=1.0\columnwidth]{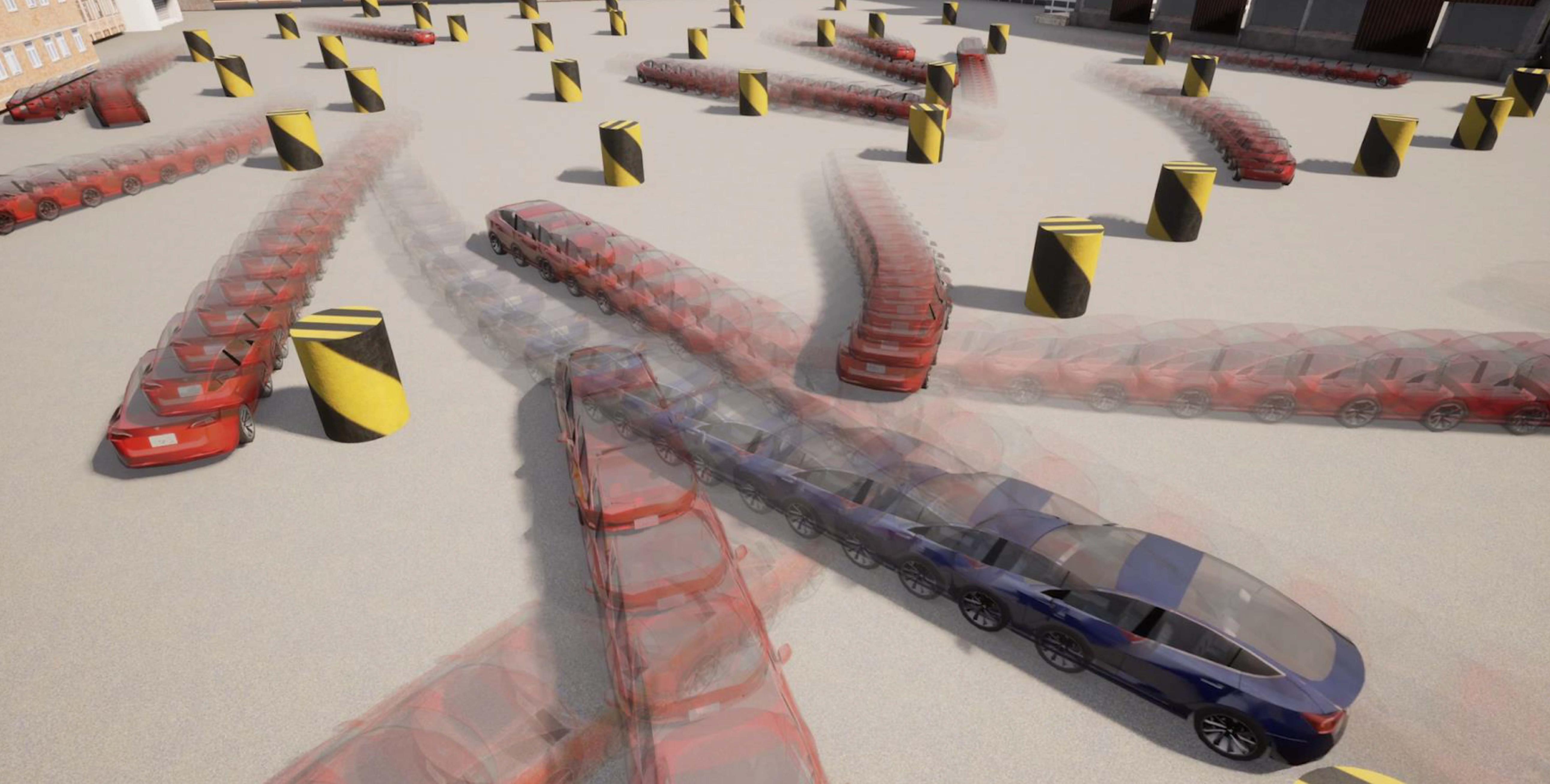}
		\caption{The movement diagram.     			}\label{fig:top1}	
		\vspace{0.1cm}
	\end{subfigure}
	\begin{subfigure}[t]{1.0\columnwidth}
		\centering
		\includegraphics[width=1.0\columnwidth]{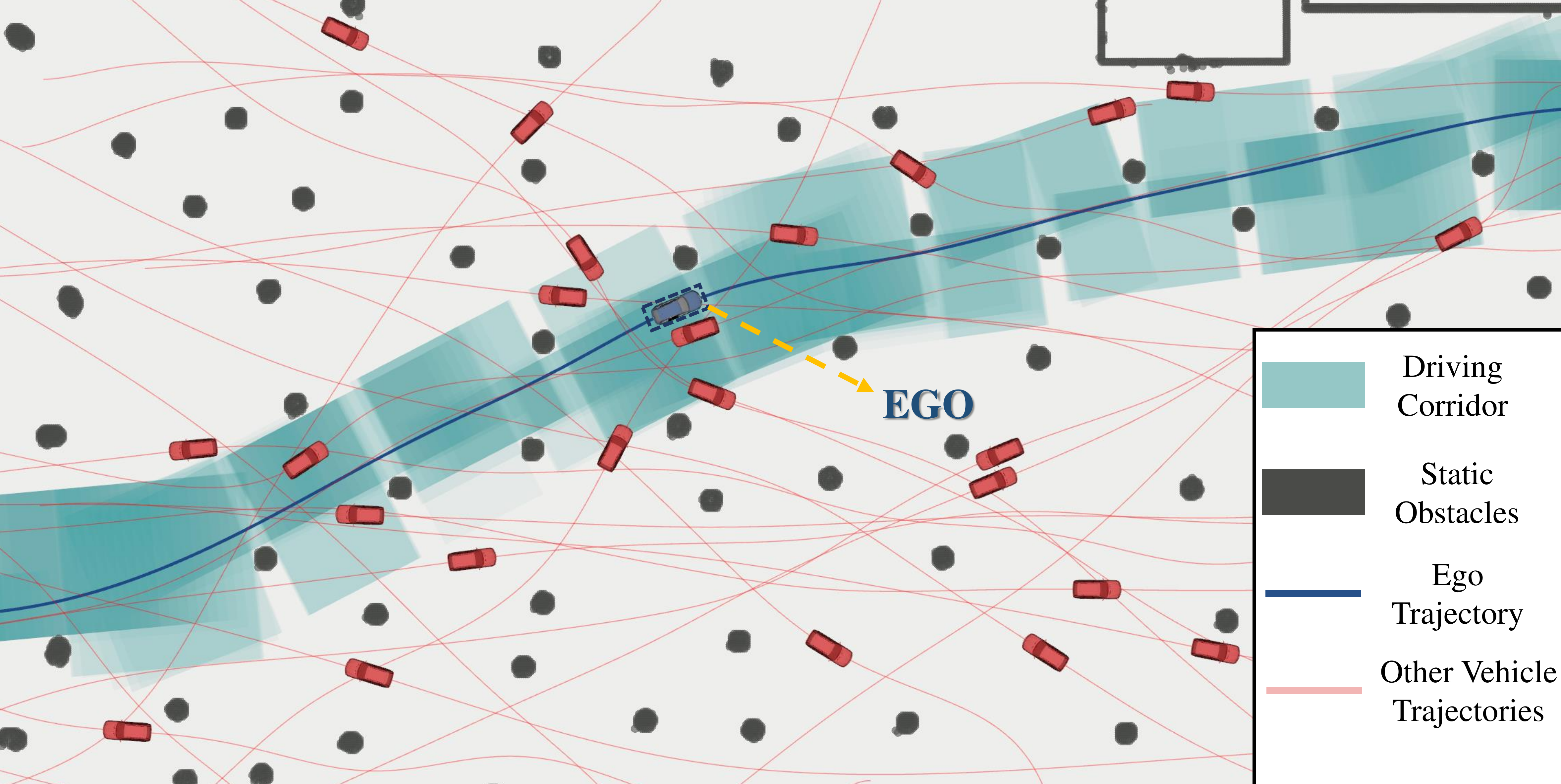}
		\caption{The trajectory visualization.
		}\label{fig:top2}
	\end{subfigure}
	\caption{The two figures show the capability of our planner in a highly dynamic environment.
		Because of full-dimensional obstacle avoidance, the ego vehicle has the ability to pass through tight areas near other moving objects while ensuring safety.}
	\label{fig:topfig}
\end{figure}

\IEEEPARstart{A}{utonomous} driving has become one of the hottest research topics in recent years because of its vast potential social benefits. It reveals a huge demand for robust and safe motion planning in complex and high-dynamic environments. Moreover, lightweight and efficiency are strongly demanded in real-world applications to ensure rapid response to dynamic and unstructured environments with limited onboard computing power. Motion planning for autonomous driving aims to generate a comfortable, low-energy, and physically feasible trajectory that makes the ego vehicle reach end states with safety guarantees and designed velocities in constrained environments. However, in recent years, only a few approaches are able to generate feasible and high-quality trajectories online in arbitrarily complex scenarios. These existing methods either rely on environmental features with predefined driving rules or oversimplify trajectory generation problems to improve time efficiency, which cannot be applied to highly constrained unstructured environments.
Overall, there is no universal solution for trajectory planning, which makes it the Achilles' heel hindering the development of autonomous driving. In fact, an ideal motion planning for autonomous driving typically faces three challenging problems.
\begin{itemize}
	\item [1)] \textbf{Nonholonomic Dynamics}: 
	Unlike holonomic robots such as omnidirectional mobile robots and quadrotors, autonomous vehicles must consider nonholonomic constraints during trajectory planning. Moreover, the strong non-convexity and nonlinearity of nonholonomic dynamics make it difficult to ensure the physical feasibility of states and control inputs in highly constrained environments.
	\item [2)] \textbf{Full-Dimensional Obstacle Avoidance}:
    Safety constraints have to balance the accuracy of object shape modeling  while maintaining an affordable computation time for online vehicle (re)planning. In real-world applications, rough approximation of the ego vehicle shape, such as one or multiple circular covering of the ego vehicle, always reduces the solution space, which introduces conservativeness or even fails to find a collision-free solution in extremely cluttered areas. While accurate modeling of the ego vehicle and other moving objects increases the complexity of the planning problem, resulting in a significant computation burden.
	
	\item [3)]\textbf{Trajectory Quality}: Time allocation is an inherent attribute of a trajectory. There is always a tradeoff between computation efficiency and trajectory quality. Common methods that optimize time and space separately reduce a large partition of solution space, especially in tightly coupled spatial-temporal scenarios such as highly dynamic environments. By contrast, spatial-temporal joint optimization can fully utilize the solution space to achieve better trajectory optimality but tends to complicate the optimization problem and reduce the real-time performance.

\end{itemize}

This paper overcomes the above critical issues by proposing an efficient spatial-temporal trajectory planning scheme that generates a  trajectory with nonholonomic constraints and achieves obstacle avoidance for full-dimensional objects, as shown in Fig. \ref{fig:topfig}. 
We parameterize the trajectory on the flat space  and encode all feasibility constraints as analytically and continuously differentiable expressions, which are integrated into an optimization formulation. Besides,  we model the geometric constraints for static obstacle avoidance and ensure dynamic safety based on the signed distance between the ego vehicle and other dynamic objects.
We conduct benchmark comparisons in simulation with other prevalent trajectory planners in different cases to demonstrate the significant superiority of the proposed method. Additionally, we validate our planner on a real platform with fully autonomous onboard computing and no external positioning.
The main contributions of this paper can be summarized as follows:
\begin{itemize}
	\item We present a continuously differentiable spatial-temporal planning formulation in which all constraint	can be analytically derived from flat outputs, thus facilitating subsequent efficient optimization.
	\item We achieve full-shape obstacle avoidance by separating static and dynamic obstacles and using a convex polygon to enclose the ego vehicle. We formulate static safety constraints based on geometric convex decomposition extracted from environments and achieve full-size dynamic avoidance with other moving obstacles by relaxing the sign distance approximation.
	\item 
	We analyze the characteristics of the constraints in the trajectory planning problem and reformulate the original optimization as an efficiently solvable one without sacrificing optimality. 
\end{itemize}


\section{Related Work}	
\label{sec:related_works}

\subsection{Trajectory Generations for Vehicles}

Sampling-based methods \cite{urmson2008autonomous,ziegler2009spatiotemporal,rufli2010design,mcnaughton2011motion,lavalle1998rapidly,shkolnik2009reachability,karaman2011sampling,webb2013kinodynamic,han2011unified,palmieri2016rrt,seegmiller2017maverick} are widely used for robot motion planning\cite{elbanhawi2014sampling} due to the ease of incorporating user-defined objectives. These typical methods, including state lattice approaches and probabilistic planners, always sample the  robot state in the configuration space to find a feasible trajectory connecting  the starting node and the goal node. Lattice-based planners\cite{urmson2008autonomous,ziegler2009spatiotemporal,rufli2010design,mcnaughton2011motion} discretize the continuous state space into a lattice graph for planning. Then, graph-search algorithms such as Dijkstra are used to obtain the  optimal trajectory in the graph.
Probabilistic planners\cite{lavalle1998rapidly,shkolnik2009reachability,karaman2011sampling,webb2013kinodynamic,han2011unified,palmieri2016rrt,seegmiller2017maverick} represented by rapidly-exploring random tree (RRT)\cite{lavalle1998rapidly} obtain a feasible path by expanding a state tree rooted at the starting node. 
Despite the ability to avoid local minima in non-convex space, sampling-based methods confront a dilemma between computation consumption and trajectory quality which limits the direct application in realistic settings.

To simplify the trajectory generation problem, some intuitive approaches\cite{zhu2015convex,liu2018convex, ZhouPJSO} decouple the spatial shape and dynamics profile of the trajectory.
Zhu et al. \cite{zhu2015convex} propose convex elastic band smoothing (CES)  algorithm which eliminates the non-convexity of the curvature constraint with fixed path lengths at each iteration, and transform the original problem into a quadratically constrained quadratic program (QCQP).
However, as presented in  work~\cite{ZhouPJSO}, the length consistency assumption does not always hold, which invalidates the curvature constraint and thus reducing the feasibility of the control.
Based on CES framework, Zhou et al. \cite{ZhouPJSO} propose the dual-loop iterative anchoring path smoothing (DL-IAPS) algorithm 
to generate a smooth and safe path, where sequential convex optimization (SCP) is used 
to relax the curvature constraint.
Whereas, the efficiency of this method relies heavily  on the initial path obtained by hybridA*\cite{dolgov2010path}, which limits the application in complex environments.

Highly adaptable model predictive control (MPC) approaches formulate a trajectory planning problem as an optimal control problem (OCP) which is further discretized into a nonlinear programming (NLP) problem. These approaches \cite{kondak2001computation,shin2018kinodynamic, li2015simultaneous,li2015unified,bergman2018combining,shi2019bilevel,zhang2018autonomous,he2021tdr,li2021optimization} directly optimize discrete states and control inputs, which can conveniently integrate dynamic and safety constraints. 
Although the above MPC-based methods can easily accommodate the motion model of robots, since the trajectory is represented by discrete states, ensuring trajectory constraints in highly constrained environments requires increasing the density of states at the expense of degraded performance.

\subsection{Obstacle Avoidance Formulation}

Modeling the ego vehicle and other obstacles determines the complexity of collision-free constraints for trajectory generation problems. Li et al. \cite{li2021optimization} cover the ego vehicle with two circles along the centerline, and then the ego vehicle model is simplified to two points by inflating obstacles. In the work\cite{li2021optimization}, obstacle avoidance is ensured by constraining these two points within a safe corridor. This approach essentially enjoys the property that the dimension of collision-free constraints is independent of the number of obstacles. However, such a point-based modeling method does not fully utilize the solution space and often generates overly conservative trajectories, especially in complex environments.

In work~\cite{li2015unified}, they considered the driving scenarios with arbitrarily placed obstacles and formulated the problem as a unified OCP for trajectory generation in unstructured environments. However, the collision-free constraint is nominally non-differentiable. Zhang et al. \cite{9062306} assume that obstacles are convex and propose an optimization-based collision avoidance (OBCA) algorithm, which removes the integer variables used to model full-dimensional object collision avoidance \cite{da2019collision}. They introduce dual variables to reformulate the distance between the robot and obstacles,  and transform the safety constraint into a continuous differentiable form. Zhang et al. \cite{zhang2018autonomous} integrate OBCA  into the MPC problem of motion planning and propose H-OBCA, a hierarchical framework for trajectory planning in unstructured environments. Besides, a reasonable method is presented in work~\cite{zhang2018autonomous} for warm-starting dual variables to speed up the optimization. However, the introduction of dual variables in OBCA-based methods\cite{zhang2018autonomous,he2021tdr}  increases the dimension of the problem, making it more challenging to solve. Besides, the number of dual variables is positively correlated with the number of obstacles. As obstacles increases, the problem dimension will rise rapidly, leading to unacceptable computational and memory costs.

Instead of sampling the planning state, decouple the spatial and temporal space, discretizing the motion model, or over-relying on the characteristics of environments, our method incorporates  spatial-temporal joint optimization to efficiently generate a high-quality trajectory with guaranteed full-dimensional obstacle avoidance. Our method enjoys low computational cost and high robustness against unstructured environments, while more detailed quantitative comparisons are presented in Sect. \ref{sec:Benchmarks}. 

\section{Spatial-Temporal Trajectory Planning}
\label{sec:Spatial-temporal Planning }
In this section, we present the spatial-temporal joint optimization formulation for trajectory planning.
To construct the optimization formulation, we discuss the complete motion planning pipeline and introduce the differential flat model of  car-like robots. Then, we show the formulation of the trajectory optimization problem in flat-output space considering human comfort, execution time, and feasibility constraints. 
Last but not least, we analyze the gradient propagation chain of the problem for subsequent numerical optimization.

\subsection{Planning Pipeline}
\label{sec:Planning Pipeline}
In practical applications, the whole pipeline follows a hierarchical structure with a front-end whose main role is to provide an initial guess, and back-end optimization. We adopt the lightweight hybridA* algorithm to find a collision-free path that is further optimized by the proposed planner.
Moreover, for each node to be expanded, we try using  Reed Shepp Curve\cite{reeds1990optimal} to shoot the end state 
for the earlier termination of the search process.
For complex driving tasks such as autonomous parking, the front-end output often contain both forward and  back vehicle movements. By reasonably assuming the vehicle always reaches a complete stop at the gear shifting position,
we parameterize the forward and backward segments of the trajectory as piece-wise polynomials, respectively, whose specific formulations are presented in Sect. \ref{sec:Optimization in Vehicle Flat-Output Space}. 
Additionally, the motion direction of each segment is determined by the front-end output and  prefixed before the back-end optimization process.

\subsection{Differentially Flat Vehicle Model}

\begin{figure}[t]  
	\vspace{-0.2cm}  
	\centering
	{\includegraphics[width=0.8\columnwidth]{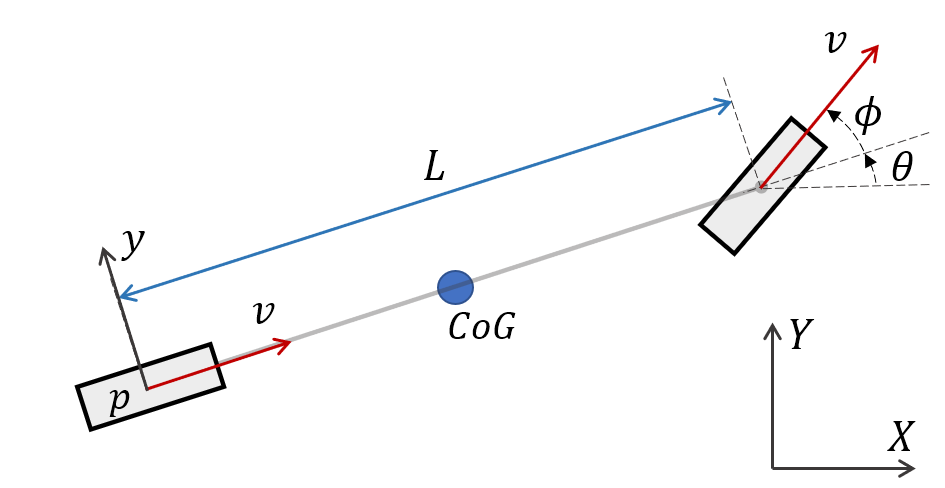}}
	\caption{ \label{fig:vehicle_model}The kinematic bicycle model.}
\end{figure}

We use the simplified kinematic bicycle model in the Cartesian coordinate frame to describe a four-wheel vehicle. Assuming that the car is front-wheel driven and steered with perfect rolling and no slipping, the model can be described as Fig. \ref{fig:vehicle_model}.  
The state vector is $\bm{x}= \left(p_x, p_y, \theta, v, a_t,a_n, \phi, \kappa\right)^{\rm T}$,  where $ \bm{p} = \left(p_x, p_y\right)^{\rm T}$ denotes the position at the center of the rear wheels, $v$ is the longitudinal velocity w.r.t vehicle's body frame, 
$a_t$ represents the longitude acceleration, $a_n$ is the latitude acceleration, $\phi$ is the steering angle of the front wheels and $\kappa$ is the curvature.  Thanks to the thorough study of the differentially flat car model \cite{Murray95differentialflatness}, we choose the flat output as $\bm{\sigma} := \left(\sigma_x, \sigma_y\right)^{\rm T}$ with a physical meaning that $ \bm{\sigma} = \bm{p} $ is the position centered on the rear wheel of the car. Other variable transformations except $p_x, p_y$  can be expressed as:
\begin{subequations}
	\begin{align}
		v     & =  \eta \sqrt{\dot{\sigma_x}^2 + \dot{\sigma_y}^2},  \\
		\theta   & =   \arctan2(\eta\dot{\sigma_y}, \eta\dot{\sigma_x}), \\
		a_t  & =  \eta (  \dot{\sigma_x}\ddot{\sigma_x}+ \dot{\sigma_y}\ddot{\sigma_y}  )   / \sqrt{\dot{\sigma_x}^2 + \dot{\sigma_y}^2} , \label{eq:at}\\
		a_n  & =  \eta (  \dot{\sigma_x}\ddot{\sigma_y}- \dot{\sigma_y}\ddot{\sigma_x}  ) /  \sqrt{\dot{\sigma_x}^2 + \dot{\sigma_y}^2 }, \label{eq:an}\\
		\phi   & =   \arctan \left( \eta (  \dot{\sigma_x}\ddot{\sigma_y}- \dot{\sigma_y}\ddot{\sigma_x}  )L /  ( \dot{\sigma_x}^2 + \dot{\sigma_y}^2 )^{\frac{3}{2}}  \right ),\\
		\kappa &= \eta (  \dot{\sigma_x}\ddot{\sigma_y}- \dot{\sigma_y}\ddot{\sigma_x}  ) /  ( \dot{\sigma_x}^2 + \dot{\sigma_y}^2 )^{\frac{3}{2}}.
	\end{align}
\end{subequations}
Consequently, with the natural differential flatness property, we can use the flat outputs and their finite derivatives to characterize arbitrary state quantities of the vehicle, which simplifies the trajectory planning and facilitates optimization. We define an additional variable $\eta \in \{-1,1\}$ to characterize the motion direction of the vehicle, where $\eta = -1$ and $\eta = 1$ represent the backward and forward movements, respectively. 
We avoid singularities by fixing the velocity magnitude  to a small, non-zero constant when the gear shifts. 
Both the gear shifting position and directional  angle can be optimized, and their specific formulations are presented in Sect. \ref{sec:Equality Constraints}.

\subsection{Optimization Formulation}
\label{sec:Optimization in Vehicle Flat-Output Space}
The $i$-th segment of the trajectory is formulated as a 2-dimensional and time-uniform $M_i$-piece polynomial with degree $N = 2s-1$, which is parameterized by the intermediate waypoints $\bm{{q}}_i = \left(\bm{{ q}}_{i,1},... , \bm{{ q}}_{i,M_i-1}\right) \in  \mathbb{R}^{2\times(M_i - 1)} $, the time interval for each piece $\delta T_i\in \mathbb{R}^+$, and the coefficient matrix $\bm{{ c}}_i = \left(\bm{{c}}_{i,1}^{\rm T} ,... , \bm{{ c}}_{i,M_i}^{\rm T} \right)^{\rm T}\in  \mathbb{R}^{2M_is\times 2}$.
Then, the $j$-th piece  of the $i$-th segment $\bm{\sigma}_{i,j}$ is written as:
\begin{equation}
	\begin{aligned}
		\bm{\sigma}_{i,j}(t) &:=  \bm{{ c}}_{i,j}^{\rm T} \bm{\beta}(t), \\
		\bm{\beta}(t) &:= \left(1, t, t^2, ..., t^N\right)^{\rm T},  \\
		\forall  t \in [0, \delta T_i], \forall i & \in \{1,2,...,n\}, \forall j \in \{1,2,3,...,M_i\}, \\
	\end{aligned}
\end{equation}
where $n$ is the number of trajectory segments and $\bm{\beta}(t)$ is a natural basis. The $M_i$-piece polynomial trajectory $\bm{\sigma}_i: [0, T_i]$ is obtained:
\begin{equation}
	\begin{aligned}
		\bm{\sigma}_i(t) &=  \bm{\sigma}_{i,j} (t-(j-1)*\delta T_i), \\
		\forall j \in \{1,2,...,&M_i\}, t \in [(j-1)*\delta T_i, j*\delta T_i).
	\end{aligned}
\end{equation}
Here, the total duration of the $i$-th segment of the trajectory
is  $T_i = M_i * \delta T_i$.
Then, the complete trajectory representation $\bm{\sigma}(t): [0,T_s]$ is formulated:
\begin{equation}
	\begin{aligned}
		\bm{\sigma}(t) &=  \bm{\sigma}_i (t-\hat T_{i}),\\
		\forall i \in \{1,2,...,&n\},  t \in [\hat T_{i}, \hat T_{i+1}),  \\
	\end{aligned}
\end{equation}
where $T_s = \sum_{i=1}^{n}T_i$ is the  duration of the whole trajectory, $\hat T_i$ = $\sum_{\bar i=1}^{i-1} T_{\bar i }$ is the timestamp of the starting point of the $i$-th segment and  $\hat T_1$ is set as 0.
Moreover, we define a coefficient  set $\bm{{\rm c}}=\left(\bm{{ c}}_1^{\rm T},\bm{{ c}}_2^{\rm T},...,\bm{{c}}_n^{\rm T}\right)^{\rm T} \in  \mathbb{R}^{\left(\sum_{i=1}^{n}2M_is\right) \times 2} $\
and a time set $\bm{{\rm T}} = \left(T_1, T_2,..., T_n\right)^{\rm T} \in \mathbb{R}^{n}$  for the subsequent derivation.
With constraints  for obstacles avoidance and dynamic feasibility, the minimal control effort problem involving time regularization  can be expressed  as a nonlinear constrained optimization:
\begin{subequations}
	\begin{align}
		\min_{\bm{{\rm c}},\bm{{\rm T}}} J({ \bm{{\rm c,T}}}) = \int_{0}^{T_s} \bm{\mu}(t)^{\rm T} \bm{ {\rm W} } \bm{\mu}(t) dt + w_TT_s   \label{eq:originopt}
	\end{align}
	\setlength\abovedisplayskip{0.1pt}
	\begin{align}
		{\rm s.t.} &\bm{\mu}(t) = \bm{\sigma}^{[s]}(t),  \forall t \in [0, T_s],  \quad \quad \quad\\		
		&\bm{\sigma}^{[s-1]}_0(0) = \bar{\bm{\sigma}}_{0}, \ \bm{\sigma}^{[s-1]}_n(T_n) = \bar{\bm{\sigma}}_{f}, \label{eq:boundary} \\
		&\bm{\sigma}^{[s-1]}_i(T_i)  = \bm{\sigma}^{[s-1]}_{i+1}(0) = \widetilde{\bm{\sigma}}_{i}, 1\leq i <n, \label{eq:gearshift} \\
		&\bm{\sigma}^{[\widetilde d]}_{i,j}(\delta T_i) = \bm{\sigma}^{[\widetilde d]}_{i,j+1}(0), 1\leq i \leq n, 1\leq j < M_i,\label{eq:continuity} \\
		&T_i > 0, 1\leq i \leq n,  \label{eq:temporal} \\
		&\mathcal{G}_d(\bm{\sigma}(t), ...,\bm{\sigma}^{(s)}(t) , t)\preceq 0, \quad \forall  d \in  \mathcal{D},  \forall t \in [0, T_s], \label{eq:user}
	\end{align}
\end{subequations}
where $\bm{ {\rm W}} \in \mathbb{R}^{2\times 2} $ is a diagonal matrix to penalize control efforts.  
Eq. (\ref{eq:boundary}) is the boundary condition, where $\bar{\bm{\sigma}}_{0}, \bar{\bm{\sigma}}_{f} \in \mathbb{R}^{2\times s}$ are the user-specified initial  and final states.
$\widetilde{\bm{\sigma}}_{i} \in \mathbb{R}^{2\times s} $ is the switching state between forward and reverse gears in which the position and the tangential direction of the motion curve are optimized. Moreover, the specific  constraint Eq. (\ref{eq:gearshift}) will be described in Sect. \ref{sec:Reformulation of  Trajectory Optimization}.
Eq. (\ref{eq:continuity}) is the continuity constraint up to degree $\widetilde d$.
The second term $ w_TT_s$ in the objective function is the time regularization term to restrict the total duration $T_s$, with a  weight $w_T \in \mathbb{R}^+$.
The constraint function at $d$ is defined as  $\mathcal{G}_d$. In our  formulation, the  set $ \mathcal{D} = \{ d : d = v, a_t,a_n, \kappa, \zeta,  \Theta \}$ includes dynamic feasibility ($v, a_t,a_n, \kappa$), static and dynamic obstacle avoidance ($\zeta, \Theta$).
Besides, $s$ is chosen as 3, which means that the integration of jerk is minimized to ensure human comfort\cite{bae2020self}.
\subsection{Gradient Derivation }
\label{sec:Gradient Derivation}
Without loss of completeness, we first derive the analytic gradients of the objective function $J$ w.r.t $ \bm{{ c }}_{i,j}$ and $T_i$:

\begin{flalign}
	&\frac{\partial J}{\partial\bm{{{ c} }}_{i,j}} = 2 \left(  \int_{0}^{\delta T_i} \bm{\beta}^{(s)}(t) \bm{\beta}^{(s)}(t)^{\rm T} dt   \right)\bm{{{ c} }}_{i,j}, \\
	&\frac{\partial J}{\partial  T_i} = \frac{1}{M_i}\sum_{j=1}^{M_i}\bm{{{ c} }}_{i,j}^{\rm T} \bm{\beta}^{(s)}(\delta T_i) \bm{\beta}^{(s)}(\delta T_i)^{\rm T}\bm{{{\rm c} }}_{i,j}+w_T.
\end{flalign}
The feasibility constraints Eq.(\ref{eq:user}) imposed on the entire trajectory are equivalent to each piece of any piece-wise polynomial trajectory segment complying with these constraints:
\begin{flalign} 
	&\mathcal{G}_d(\bm{\sigma}(t), ...,\bm{\sigma}^{(s)}(t) , t)\preceq {0}, \forall  d \in  \mathcal{D},  \forall t \in [0, T_s] \iff \nonumber\\
	&\mathcal{G}_d(\bm{\sigma}_{i,j}(\bar t), ...,\bm{\sigma}^{(s)}_{i,j}(\bar t) , \hat t)\preceq {0},\forall  d \in  \mathcal{D},  \forall \bar t \in [0, \delta T_i],	 \nonumber \\
	& \quad \quad  \forall i  \in \{1,2,...,n\}, \forall j \in \{1,2,3,...,M_i\},\label{eq:piececonstraint}
\end{flalign}
where $\bar t$ is the relative timestamp and $\hat t = \hat T_i + \delta T_i*(j-1)+\bar t $ is the absolute timestamp. 
Therefore, to approximate the continuous-time formula Eq. (\ref{eq:piececonstraint}), we uniformly discretize each piece of the piece-wise polynomial into $\lambda \in \mathbb{N}_{>0}$ constraint points.                   
Moreover, we ensure trajectory feasibility by imposing constraints on these constraint points. 
Then,  the continuous-time formula Eq. (\ref{eq:piececonstraint}) is transformed into a discrete form:
\begin{flalign} 
	&\mathcal{G}_{d,i,j,k}(\bm{{{ c} }}_{i,j}, \bm{\mathrm{T}}) \preceq 0,  \nonumber\\
	&\mathcal{G}_{d,i,j,k}(\bm{{{ c} }}_{i,j}, \bm{\mathrm{T}}) := \mathcal{G}_d(\bm{\sigma}_{i,j,k}, ...,\bm{\sigma}^{(s)}_{i,j,k},\hat t), \nonumber\\
	&\bm{\sigma}^{(\bar d)}_{i,j,k} := \bm{\sigma}_{i,j}^{(\bar d)}(\bar t), \quad \forall \bar d \in \{0,1,...,s\},\nonumber\\
	&\bar t = \frac{k T_i}{\lambda M_i}, \quad \hat t = \hat T_i + \left(\frac{j-1}{M_i}+\frac{k}{\lambda M_i}\right)T_i, \nonumber\\
	&\forall k \in  \{0,1,2,..., \lambda \}, \quad \forall  d \in  \mathcal{D}. \label{eq:pointconstraint} 
\end{flalign}Without loss of generality, we derive the  gradient propagation at any constraint point based on the chain rule:
\begin{flalign}
	\frac{\partial \mathcal{G}_{d,i,j,k}}{\partial {\bm{{ c }}}_{i,j} } &= 
	\sum_{\bar d = 0}^{s}\beta^{(\bar d)}(\bar t)
	{\left(\frac{\partial \mathcal{G}_{d,i,j,k}}{\partial {\bm{{\rm \sigma}}_{i,j,k}^{(\bar d)}}}
		\right)}^{\rm T},\label{eq:gradGtoc}\\
	\frac{\partial \mathcal{G}_{d,i,j,k}}{\partial  {\bm{\mathrm{T}}}}	&= 
	\frac{\partial \mathcal{G}_{d,i,j,k}}{\partial \bar t}
	\frac{\partial \bar t}{\partial {\bm{\mathrm{T}}}}
	+
	\frac{\partial \mathcal{G}_{d,i,j,k}}{\partial \hat t}
	\frac{\partial \hat t}{\partial {\bm{\mathrm{T}}}},\label{eq:gradGtoT} 
\end{flalign}We further derive  time-related gradient terms inside Eq. (\ref{eq:gradGtoT}):
\begin{flalign}
	\frac{\partial \mathcal{G}_{d,i,j,k}}{\partial \bar t}&=
	\sum_{\bar d = 0}^{s} 
	{\left({{\bm{\mathrm{\sigma}}}_{i,j,k}^{(\bar d+1)}}\right)}^{\rm T}
	{\frac{\partial \mathcal{G}_{d,i,j,k}}{\partial {\bm{{\rm \sigma}}_{i,j,k}^{(\bar d)}}}},
	\label{eq:Gtobart}
	\\
	\frac{\partial \bar t}{\partial \bm{\mathrm{T}}}&=
	\left(\mathbf{0}_{i-1}^{\rm T},
	\frac{k}{\lambda M_i},
	\mathbf{0}_{n-i}^{\rm T}\right)^{\rm T},\label{eq:gradbarttoT}
	\\  
	\frac{\partial \hat t}{\partial \bm{\mathrm{T}}}&=
	\left(
	\mathbf{1}_{i-1}^{\rm T},
	\frac{k}{\lambda M_i}+\frac{j-1}{M_i},
	\mathbf{0}_{n-i}^{\rm T}
	\right)^{\rm T},\label{eq:gradhattoT}
\end{flalign}
where  all the above gradient calculations w.r.t vectors follow the denominator layout.
As a result, by substituting Eq.(\ref{eq:Gtobart})-(\ref{eq:gradhattoT}) to
Eq.(\ref{eq:gradGtoT}), we can obtain the gradients of $G_{d,i,j,k}$ w.r.t $\bm{{c}}_{i,j}$ and $\bm{\mathrm{T}}$
once  $\partial\mathcal{G}_{d,i,j,k}/{{\partial {\bm{{\rm \sigma}}_{i,j,k}^{(\bar d)}}}}$ and 
$\partial \mathcal{G}_{d,i,j,k}/{\partial \hat t}$ are specified. 
Besides, it is worth mentioning that  constraint  functions $G_{d,i,j,k}$ 
can be precisely expressed  by some of the quantities in $\hat t$ and ${\bm{{\rm \sigma}}_{i,j,k}^{(\bar d)}}$, where $\bar d\in \{0,1,2...,s\}$.
Therefore, the gradients w.r.t irrelevant variables are 0 without derivation.
In subsequent sections, we  present the specific formulation of the constraint functions $\mathcal{G}_{d\in \mathcal{D} }$ and derive gradients.
For simplification, $i,j,k$ and relative timestamp $\bar t$ are omitted in Sect. \ref{sec:Static States Constraints} and Sect. \ref{sec:Dynamic Avoidance}.
\section{Instantaneous State Constraints}
\label{sec:Static States Constraints}
In this section, we introduce the  instantaneous state constraints for trajectory optimization, where these constraint functions are only related to  the  instantaneous states of the vehicle.
\subsection{Dynamic Feasibility}
\label{sec:Dynamic Feasibility}
\subsubsection{Longitude  Velocity Limit}
For autonomous driving, the longitude velocity always needs to be limited within a reasonable range because of practical factors such as traffic rules, physical vehicle performance, and environmental uncertainty.
Then, the constraint  function of longitude velocity  at a  constraint point is defined as follows:
\begin{flalign} \label{eq:va_const}
	&\mathcal{G}_{v}(\dot{\bm{\sigma}}) = \dot{\bm{\sigma}}^{T}\dot{\bm{\sigma}} - v_m^2.
\end{flalign}
where $v_m$ is the magnitude of maximal longitude velocity.
The gradient of $\mathcal{G}_{v}(\dot{\bm{\sigma}})$ is written as:
\begin{equation}
	\frac{\partial \mathcal{G}_{v}}{\partial \dot{\bm{{\rm \sigma}}}} = 2 {\dot{\bm{\sigma}}}.\label{eq:gvtosigma}
\end{equation}
Accordingly,  we can obtain the gradients of $\mathcal{G}_{v}$ by combining Eq.(\ref{eq:gvtosigma}) and Eq.(\ref{eq:gradGtoc})-(\ref{eq:gradhattoT}).
\subsubsection{Acceleration Limit}
The  acceleration  is always required to be limited to prevent skidding due to  friction limits between the tire and the ground.
From Eq. (\ref{eq:at})(\ref{eq:an}), we define the constraint  functions of longitude  and latitude acceleration  at a  constraint point:
\begin{flalign} 
	&\mathcal{G}_{a_t}(\dot{\bm{\sigma}}, \ddot{\bm{\sigma}}) =  \frac{(\ddot{\bm{\sigma}}^T\dot{\bm{\sigma}})^2}{{\dot{\bm{\sigma}}}^{T}{\dot{\bm{\sigma}}} } -  a_{tm}^2,\\
	&\mathcal{G}_{a_n}(\dot{\bm{\sigma}}, \ddot{\bm{\sigma}}) =  \frac{(\ddot{\bm{\sigma}}^T \bm{{\rm B}} \dot{\bm{\sigma}})^2}{{\dot{\bm{\sigma}}}^{T}{\dot{\bm{\sigma}}} } -  a_{nm}^2,
\end{flalign}
where $a_{tm}$ and $a_{nm}$ are the  maximal longitude and latitude acceleration and  $ \bm{{\rm B}} := \begin{bmatrix}
	0 & -1\\1 & 0\end{bmatrix} $ is an auxiliary antisymmetric  matrix.
The gradients of $\mathcal{G}_{a_t}(\dot{\bm{\sigma}}, \ddot{\bm{\sigma}})$ and $\mathcal{G}_{a_n}(\dot{\bm{\sigma}}, \ddot{\bm{\sigma}})$ are derived as :
\begin{flalign} 
	& \frac{\partial \mathcal{G}_{a_t}}{\partial \dot{\bm{{\rm \sigma}}}} = 
	2\frac{\ddot{\bm{\mathrm{\sigma}}}^{T}\dot{\bm{\mathrm{\sigma}}} }{||\dot{\bm{\mathrm{\sigma}}}||_2^2}\ddot{\bm{\mathrm{\sigma}}} - 
	2\left( 
	\frac{\ddot{\bm{\mathrm{\sigma}}}^{T}\dot{\bm{\mathrm{\sigma}}} }{||\dot{\bm{\mathrm{\sigma}}}||_2^2}
	\right)^2\dot{\bm{\mathrm{\sigma}}}
	, \label{eq:gatodsigma}\\
	& \frac{\partial \mathcal{G}_{a_n}}{\partial \dot{\bm{{\rm \sigma}}}} = 
	2\frac{\ddot{\bm{\mathrm{\sigma}}}^{T} \bm{{\rm B}} \dot{\bm{\mathrm{\sigma}}} }{||\dot{\bm{\mathrm{\sigma}}}||_2^2} \bm{{\rm B}}^T\ddot{\bm{\mathrm{\sigma}}} - 
	2\left( 
	\frac{\ddot{\bm{\mathrm{\sigma}}}^{T} \bm{{\rm B}}\dot{\bm{\mathrm{\sigma}}} }{||\dot{\bm{\mathrm{\sigma}}}||_2^2}
	\right)^2\dot{\bm{\mathrm{\sigma}}}
	, \label{eq:gantodsigma}\\
	& \frac{\partial \mathcal{G}_{a_t}}{\partial \ddot{\bm{{\rm \sigma}}}} = 
	2\frac{\ddot{\bm{\mathrm{\sigma}}}^{T}\dot{\bm{\mathrm{\sigma}}}}
	{||\dot{\bm{\mathrm{\sigma}}}||_2^2}\dot{\bm{\mathrm{\sigma}}},	\frac{\partial \mathcal{G}_{a_n}}{\partial \ddot{\bm{{\rm \sigma}}}} = 
	2\frac{\ddot{\bm{\mathrm{\sigma}}}^{T} \bm{{\rm B}} \dot{\bm{\mathrm{\sigma}}}}
	{||\dot{\bm{\mathrm{\sigma}}}||_2^2}\bm{{\rm B}}\dot{\bm{\mathrm{\sigma}}}
	\label{eq:gatoddsigma}.
\end{flalign}

\subsubsection{Front Steer Angle Limit}
The front steer angle needs to be limited to ensure the nonholonomic dynamic feasibility of the vehicle.
Due to the monotonicity of the tangent function, we restrict the front steer angle by limiting the curvature $\kappa = \tan \phi/L$ in 
$[ -\tan \phi_{m}/L, \tan \phi_{m}/L ] := [-\kappa_{m}, \kappa_{m}]$, where $\phi_{m}$  is the preset maximum steer angle
and $\kappa_{m}$ is the corresponding maximum curvature.
Then, the nonholonomic dynamic constraint  function $\mathcal{G}_{{\kappa}}$ is 
expressed  as:
\begin{flalign} \label{eq:k_const}
\mathcal{G}_{\kappa}(\dot{\bm{\sigma}},\ddot{\bm{\sigma}}) = (\frac{\ddot{\bm{\sigma}}^T \bm{{\rm B}} \dot{\bm{\sigma}}   }{ {||\dot{\bm{\sigma}}||}_2^3})^2 - \kappa_{m}^2.
\end{flalign}
Furthermore, we derive the gradients w.r.t $\dot{\bm{\sigma}}$ and $\ddot{\bm{\sigma}}$:
\begin{flalign}
	& \frac{\partial \mathcal{G}_{\kappa}}{\partial \dot{\bm{{\rm \sigma}}}} = 
2(\frac{\ddot{\bm{\sigma}}^T \bm{{\rm B}} \dot{\bm{\sigma}}   }{ {||\dot{\bm{\sigma}}||}_2^3})(\frac{\bm{{\rm B}}^T \ddot{\bm{\sigma}}}{ {||\dot{\bm{\sigma}}||}_2^3} - 
	3\frac{\ddot{\bm{\sigma}}^T \bm{{\rm B}} \dot{\bm{\sigma}}}{{||\dot{\bm{\sigma}}||}_2^5}\dot{\bm{\sigma}}), \\
	& \frac{\partial \mathcal{G}_{\kappa}}{\partial \ddot{\bm{{\rm \sigma}}}} =
	2(\ddot{\bm{\sigma}}^T \bm{{\rm B}} \dot{\bm{\sigma}}   )\frac{\bm{{\rm B}} \dot{\bm{\sigma}}}{{||\dot{\bm{\sigma}}||}_2^6}.
\end{flalign}
\subsection{Static Obstacle Avoidance}
\label{sec:Static Obstacle Avoidance Contraints}
\begin{figure}[t]  
	\vspace{-0.0cm}  
	\centering
	{\includegraphics[width=1.0\columnwidth]{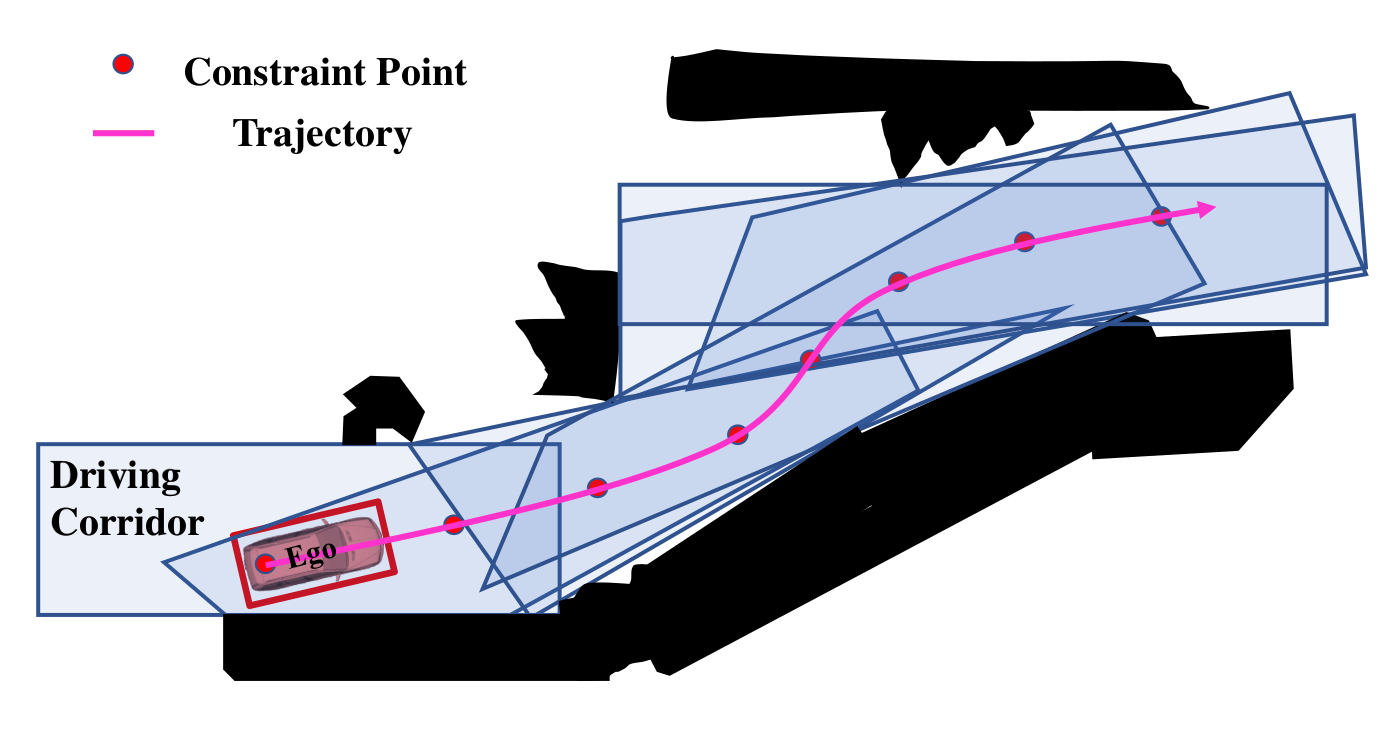}}
	\caption{ \label{fig:obstacle}Visualization of the safe driving corridor. The convex polygon contains the full shape of the ego vehicle at any constraint point.
		With a proper trajectory resolution, we can practically guarantee the static no-collision property.}
\end{figure}

In this subsection, we analytically present static safety constraints that are efficiently computable based on the geometric representation of the 	free space in  the environment.
We first decompose the semantic environment and extract the safe space to construct a driving corridor consisting of a series of convex polygons.
Then, we derive the necessary and sufficient condition for enforcing the full vehicle shape in the driving corridor, which is used to construct  static no-collision constraints.
Before  specific derivation, we introduce the pipeline of the constraint modeling.
We first discretize the collision-free path generated by the front end into sampling points whose number is the same as the number of constraint points in the back-end optimization.
Then, combined with the environmental information, we generate a free convex polygon based on the sampling point by the method \cite{Zhong2020GeneratingLC} or directly expanding in each defined direction. As a result, the entire trajectory is guaranteed to be safe by confining the full vehicle shape at each constraint point to the corresponding convex polygon, as shown in Fig. \ref{fig:obstacle}.
We use a convex polygon to enclose the full shape of the ego vehicle which is defined as $\mathbb{E}$.
Moreover, we  define the vertice set $\mathcal{E}$ of the convex polygon  as:
\begin{equation}
	\mathcal{E} =  \{ \bm{v}_e \in \mathbb{R}^2  : \bm{v}_e  = \bm{\sigma} +  \bm{{\rm R}} \bm{l}_e, e=1,2,...,n_e \}, \label{eq:Evertics}
\end{equation} 
where   $\bm{{\rm R}}$ is the rotation matrix from the body  to the world frame, transformed into flat outputs as: 
\begin{equation}
	\bm{{\rm R}} =  \frac{\eta}{\left \| \dot{\bm{\sigma}} \right \|_2} \left(\dot{\bm{\sigma}} ,\bm{{\rm B}} \dot{\bm{\sigma}} \right).
\end{equation}
Here, $n_e$ is the number of vertexes, and $\bm{l}_e$ is the coordinate of the $e$-th vertex in the body frame. 
$\eta\in\{-1,1\}$ is a prefixed auxiliary variable to indicate the motion direction of the segment of the trajectory.
Note $n_e$ and  $\bm{l}_e$ are also constant once the vehicle shape is identified.
The H-representation~\cite{avis2002canonical} of each convex polygon $\mathcal{P}^H$ in the driving corridor is  obtained:
\begin{equation}
	\begin{aligned}
		\mathcal{P}^H &=  \{ \bm{q}\in \mathbb{R}^2:  \bm{A} \bm{q} \leq \bm{b} \},\\
		\bm{A} &=\left( \bm{A}_1, ..., \bm{A}_z, ..., \bm{A}_{n_{z}} \right)^{\rm T} \in \mathbb{R}^{n_{z}\times 2} ,\\ 
		\bm{b} &= \left(b_1, ..., b_z, ..., b_{n_z} \right)^{\rm T} \in \mathbb{R}^{n_{z}}, \\
	\end{aligned}
\end{equation}
where $n_z$ is the number of hyperplanes, 
$\bm{A}_z \in \mathbb{R}^{2 }$ and ${b}_z \in \mathbb{R}$ are the descriptors of a hyperplane, which can be determined by a point on the hyperplane and the normal vector.  Additionally, once the driving corridor is generated, the hyperplane descriptors $\bm{A}_z$ and ${b}_z$
are also  completely determined.
A sufficient and necessary condition for containing the full shape of the vehicle in a convex polygon is that each vertex of the vehicle shape is contained in the convex polygon:
\begin{equation}
	\begin{aligned}
		\mathbb{E} \subseteq  \mathcal{P}^H &\iff \\
		\bm{\sigma} + &\bm{{\rm R}} \bm{l}_e \subseteq  \mathcal{P}^H \quad \forall e \in \{1,2,...,n_e\}.
	\end{aligned}
\end{equation}
A sufficient and necessary condition for containing a vertex  in a convex polygon is that the vertex is on the inner side of each hyperplane:
\begin{equation}
	\begin{aligned}
		\bm{\sigma} + &\bm{{\rm R}} \bm{l}_e \subseteq  \mathcal{P}^H \iff \\
		&\bm{A}_z^{\rm T} \left(\bm{\sigma} + \bm{{\rm R}} \bm{l}_e \right) \leq b_z \quad \forall z \in \{1,2,...,n_{z}\}.
	\end{aligned}
\end{equation}
Therefore, the spatial constraint function at a constraint point is $\mathcal{G}_{{\zeta}} = (\mathcal{G}_{\zeta_{1,1}}, ..., \mathcal{G}_{\zeta_{e,z}}, ..., \mathcal{G}_{\zeta_{n_e,n_z}})^{\rm T} \in \mathbb{R}^{n_en_z}$, with $n_en_z$ linear constraint penalty about  vertices of the ego vehicle, which is defined as:
\begin{equation}
	\mathcal{G}_{\zeta_{e,z}}(\bm{\sigma},\dot{\bm{\sigma}} ) = \bm{A}_z^{\rm T} (\bm{\sigma}+ \bm{{\rm R}}\bm{l}_e) - b_z. 
\end{equation}
Before further derivation, we define an auxiliary expression $\mathcal{F}(\bm{l}):\mathbb{R}^2\rightarrow\mathbb{R}^{2\times2}$ to simplify the form:
\begin{flalign}
	\mathcal{F}({\bm{l}}) =  \frac{ \eta \left(\bm{l},  \bm{{\rm B}}\bm{l}\right)^{\rm T} }{\left \| \dot{\bm{\sigma}} \right \|_2} -
	\frac{ \dot{\bm{\sigma}} \left(\bm{\mathrm{R}} \bm{l}\right)^{\rm T}  }{ \left \| \dot{\bm{\sigma}} \right \|_2^2}  .
\end{flalign}	
The gradients of the constraint function  w.r.t $\bm{\sigma}$ and $\dot{\bm{\sigma}}$ are:
\begin{flalign}
	\frac{\partial\mathcal{G}_{\zeta_{e,z}}}{\partial \bm{\sigma} } & =  \bm{A}_{z}, \\
	\frac{\partial\mathcal{G}_{\zeta_{e,z}}}{\partial \dot{\bm{\sigma}} } & =  \mathcal{F}({\bm{l}}_e)\bm{A}_z .
\end{flalign}	
The gradients w.r.t the polynomial coefficients and durations  can also be calculated by propagating equations  Eq.(\ref{eq:gradGtoc})-(\ref{eq:gradhattoT}).
\section{Dynamic Obstacle Avoidance}
\label{sec:Dynamic Avoidance}
Dynamic safety is guaranteed by ensuring that the minimum distance between the ego vehicle  and  obstacle  convex polygons at each moment of the trajectory is greater than the safety threshold. 
To increase readability, we introduce  helpful priors for evaluating  the signed distance between convex polygons.
Then, we elaborate on the dynamic avoidance constraint function which  is further relaxed into a continuously differentiable form.
\subsection{Preliminaries on Distance Representations}
\label{sec:preliminaries}
\subsubsection{Signed Distances for Rigid Objects}
We consider two convex polytopes $\mathbb{E}, \mathbb{O}  $ bounded by the intersection
of halfspaces, as $\mathbb{E} = \cap_{e=1}^{K_{e}} \mathcal{P}_e$, $\mathbb{O} = \cap_{o=1}^{K_{o}} \mathcal{P}_o$. 
The usual convex collision avoidance method penalizes the signed distance between two sets.
The distance is defined with the minimal translation $\mathcal{T}$ as:
\begin{equation}
	{\rm dist}(\mathbb{E}, \mathbb{O}) =  \min_{\mathcal{T}}\{\left \| \mathcal{T}\right \|: (\mathbb{E} + \mathcal{T})\cap O \neq \O\}.\\
	\label{eq:dist}
\end{equation}
When overlapping, ${\rm dist} = 0 $ holds, which is  insufficient to obtain gradient directions to separate them. The penetration depth can be combined to solve this issue:
\begin{equation}
	{\rm pen}(\mathbb{E}, \mathbb{O}) =  \min_{\mathcal{T}}\{\left \|\mathcal{T}\right \|: (\mathbb{E} + \mathcal{T})\cap O = \O\}.\\
	\label{eq: pen}
\end{equation}
Hence, we can get the signed distance:
\begin{equation}
	{\rm sd}(\mathbb{E}, \mathbb{O}) := {\rm dist}(\mathbb{E}, \mathbb{O}) - {\rm pen}(\mathbb{E}, \mathbb{O}).\\
	\label{eq:sd}
\end{equation}
Computing the signed distance requires solving  minimum optimization problems Eq. (\ref{eq:dist}, \ref{eq: pen}), which is unsuitable for embedding into our trajectory optimization problem.
We will discuss a Minkowski difference-based algorithm for the approximate efficient computation of signed distances in the next section.
\subsubsection{Approximation Distances}
\label{sec:Bounds}
\begin{figure}[t]   
	\centering
	{\includegraphics[width=0.95\columnwidth]{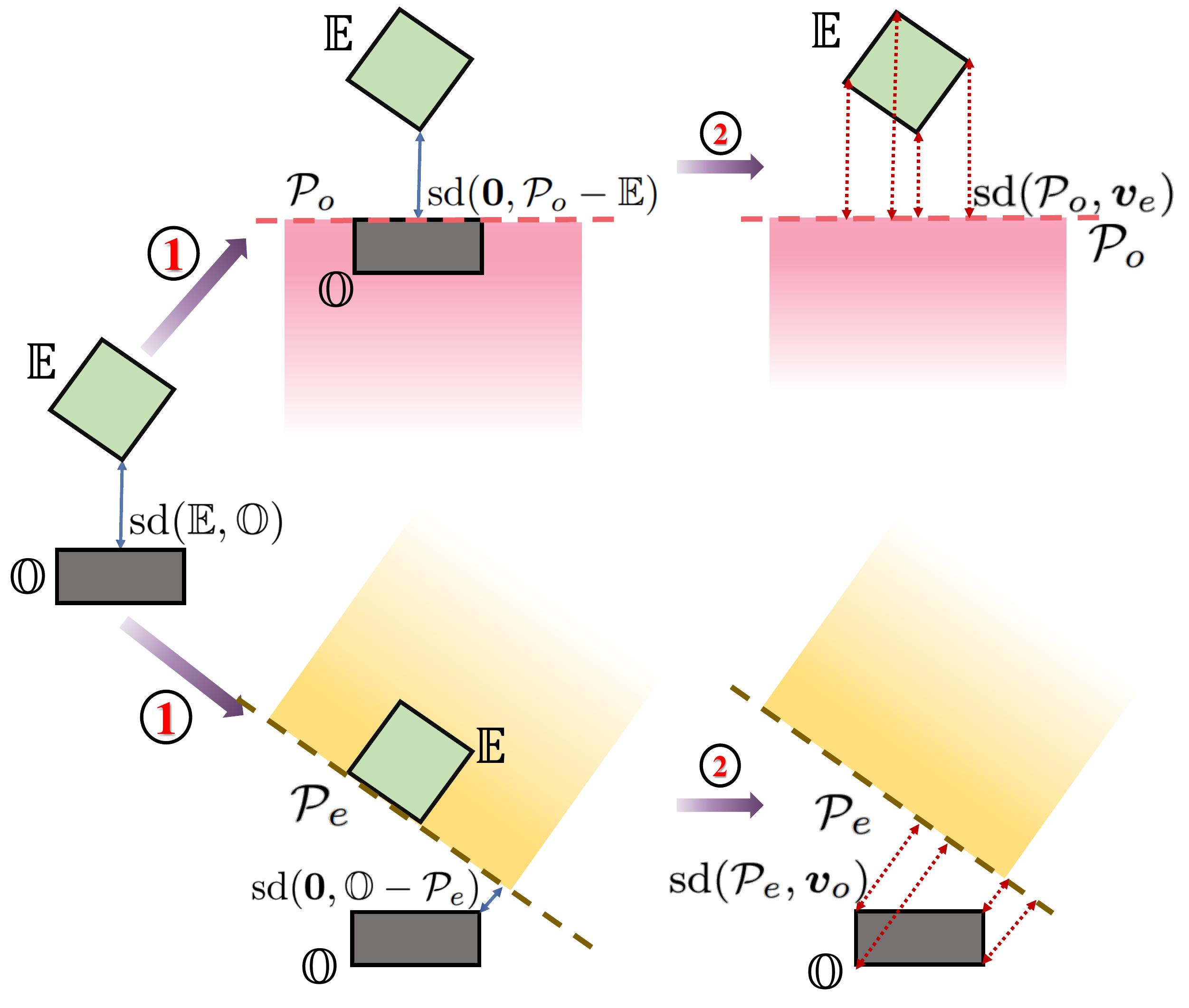}}
	\caption{ \label{fig:dynpre}Illustration of computing the lower bound of the signed distance between  convex sets $\mathbb{E}$ and $\mathbb{O}$,
		where $\mathcal{P}_e$ and $\mathcal{P}_o$ refer to any hyperplane of $\mathbb{E}$ and $\mathbb{O}$.
		Besides, we define $\bm{v}_e$ and $\bm{v}_o$ as any vertex of  $\mathbb{E}$ and $\mathbb{O}$, respectively.
		The computation process follows the above two-stage structure.
		We first use the maximum of the signed distances between convex sets and hyperplanes to
		approximate ${\rm sd}(\mathbb{E}, \mathbb{O})$. 
		Then, due to the convexity of $\mathbb{E}$ and $\mathbb{O}$, ${\rm sd}(\bm{0},\mathcal{P}_o-\mathbb{E})$ and ${\rm sd}(\bm{0},\mathbb{O}-\mathcal{P}_e)$ are converted to point-to-hyperplane distances
		${\rm sd}({\mathcal{P}_o,\bm{v}_e})$ and ${\rm sd}({\mathcal{P}_e,\bm{v}_o})$ which can be analytically calculated.
	}
	\vspace{-0.0cm} 
\end{figure}

We follow the definition of Minkowski difference used in GJK algorithm\cite{2083}. Considering the general case where $A, B \in  \mathbb{R}^n $ are two sets, the Minkowski Difference is defined by: 
\begin{equation}
	A-B = \{a-b\in \mathbb{R}^n: a \in A, b \in B  \}.\\
\end{equation}Based on the core property \cite{1087645} extensively used for collision checking:
\begin{flalign}
	{\rm sd}( \mathbb{E}, \mathbb{O}) = {\rm sd}(\bm{0}, \mathbb{O}- \mathbb{E}).
\end{flalign}
The problem of computing signed distances between two sets can be reduced to the distance between the origin point $\bm{0}$  to the set $\mathbb{O}- \mathbb{E}$. A concise formulation to bound signed distances is proposed in \cite{lutz2021efficient}, defined as the maximum signed distance from the origin to the Minkowski difference between a polygon and each hyperplane:
\begin{flalign}
	\max_{\mathcal{P}_e, \mathcal{P}_o} \{{\rm sd}(\bm{0}, \mathbb{O}- \mathcal{P}_e), {\rm sd}(\bm{0},\mathcal{P}_o-\mathbb{E})\} \leq  {\rm sd}(\mathbb{E}, \mathbb{O}) , \notag \\
	\forall  e = \{1,..., K_{e}\},\forall  o = \{1,..., K_{o}\}.
	\label{eq:lower_bound}
\end{flalign}By extending the Minkowski difference to the case between a polygon and each hyperplane, we have
\begin{flalign}
	\mathbb{O}- \mathcal{P}_e &= \{\bm p\in \mathbb{R}^n: \bm p + \bm y  \in \mathbb{O}, \bm y \in \mathcal{P}_e  \},  \notag \\
	&= \{\bm p \in \mathbb{R}^n  : {\left(\bm{{\rm H}}^e\right)}^{\rm T}\bm p \geq  -{\rm h}^e +  {\left(\bm{{\rm H}}^e\right)}^{\rm T}\bm u, \  \bm u \in \mathbb{O}\}.
\end{flalign}
with $ \mathcal{P}_e = \{\bm y \in \mathbb{R}^n:  {\left(\bm{{\rm H}}^e\right)}^{\rm T} \bm y \leq  {\rm h}^e  \} $ as a hyperplane of the set $\mathbb{E}$. Similarly, we obtain
\begin{flalign}
	\mathcal{P}_o - \mathbb{E} &= \{\bm p \in \mathbb{R}^n  : {\left(\bm{{\rm G}}^o\right)}^{\rm T} \bm p \leq  {\rm g}^o -  {\left(\bm{{\rm G}}^o\right)}^{\rm T} \bm y, \  \bm y \in \mathbb{E}\}.
\end{flalign}
The signed distance can be computed as:
\begin{flalign}
	{\rm sd}(\bm{0}, \mathbb{O}- \mathcal{P}_e) &= \frac{1}{\left \| \bm{{\rm H}}^e \right \|_2}(-{\rm h}^e + \min_{\bm u} {\left(\bm{{\rm H}}^e\right)}^{\rm T}\bm u ) ,\label{eq:disOpe}\\
	{\rm sd}(\bm{0}, \mathcal{P}_o - \mathbb{E}) &= \frac{1}{\left \| \bm{{\rm G}}^o \right \|_2}(-{\rm g}^o + \min_{\bm y}  {\left(\bm{{\rm G}}^o\right)}^{\rm T}\bm y ) .
\end{flalign}
The physical meaning of this formulation is that the signed distance of two sets is bounded by a maximum signed distance of one set to each hyperplane and vice versa. For convex polytopes, we only need to check each vertex to get the minimum value 
and then the lower bound can be analytically calculated for optimization, as shown in Fig. \ref{fig:dynpre}.
\subsection{Constraint  for Dynamic Avoidance}
\label{sec:Constraint Violation for Dynamic Avoidance}
We apply discrete obstacle avoidance constraints on  the constraint points along the trajectory regarding the trajectories of other obstacles at the same time stamp. 
Therefore, the dynamic safety constraint  is defined as $\mathcal{G}_ \Theta(\bm{\sigma},\dot{{\bm{\sigma}}},\hat t) = \{\mathcal{G}_{ \Theta_1}, ..., \mathcal{G}_{ \Theta_u}, ..., \mathcal{G}_{ \Theta_{N_u}}\}^{\rm T} \in 
\mathbb{R}^{N_u}$ where $N_u$ is the number of dynamic obstacles. The dynamic avoidance constraint function with the $u$-th moving object at a constraint point is defined as:
\begin{flalign}
	\mathcal{G}_{ \Theta_u}(\bm{\sigma},\dot{\bm{\sigma}}, \hat{t}) = d_{m} - U(\mathbb{E}(\bm{\sigma},\dot{\bm{\sigma}}), \mathbb{O}_{u}(\hat{t})) ),\label{eq:dviolation}
\end{flalign}
where $ d_{\rm {m}} $ is the minimum safe distance (safety margin) and $ U(\mathbb{E}(t), \mathbb{O}_{u}(\hat{t}))$ is the  distance between the ego vehicle and the moving obstacle. 
With the lower approximation of the signed distance between two convex objects (Sect. \ref{sec:Bounds}), we have:
\begin{flalign}
	{\rm sd }\left(\mathbb{E}(\bm{\sigma},\dot{\bm{\sigma}}\right), \mathbb{O}_{u}(\hat{t})) &\geq  {\rm lb_{sd} }\left(\mathbb{E}(\bm{\sigma},\dot{\bm{\sigma}}\right), \mathbb{O}_{u}(\hat{t})).
\end{flalign}
The lower bound ${\rm lb_{sd}}$ is paraphrased by Eq. (\ref{eq:lower_bound}) as:
\begin{flalign}
	&{\rm lb_{sd} }\left(\mathbb{E}(\bm{\sigma},\dot{\bm{\sigma}}), \mathbb{O}_{u}(\hat t)\right) = \nonumber\\
	&\max_{\mathcal{P}_e,\mathcal{P}^u_o} \left \{ 
	{\rm sd}(\bm{0},\mathbb{O}_{u}(\hat t)-\mathcal{P}_e(\bm{\sigma},\dot{\bm{\sigma}})), {\rm sd}(\bm{0},\mathcal{P}_{o}^u(\hat t)-\mathbb{E}(\bm{\sigma},\dot{\bm{\sigma}}))
	\right \},\nonumber\\
	& \quad \quad e \in \{1,...,n_e\}, o\in\{1,...,n_u\} \label{eq:lowerbound},
\end{flalign}where $n_u$ is the number of  hyperplanes of the $u$-th  moving obstacle and $\mathcal{P}_{o}^u$ is the hyperplane.
Since the motion planning is on a two-dimensional plane, a hyperplane of a convex set degenerates into a straight line determined by two vertices.
Then, the hyperplane descriptors of the ego vehicle can be determined as follows:
\begin{flalign}
	\bm{{\rm H}}^e &=\frac{\bm{{\rm B}} \left(\bm{v}_{e+1}-\bm{v}_{e}\right)}{||\bm{v}_{e+1}-\bm{v}_{e}||_2}, \label{eq:He}\\
	{\rm h}^e & = {\left(\bm{{\rm H}}^e\right)}^{\rm T} \bm{v}_e, e \in \{1,...,n_e\} \label{eq:he},
\end{flalign} 
where $\{\bm{v}_{1}, ..., \bm{v}_{e}, ..., \bm{v}_{n_e+1}\}$ are the vertices arranged clockwise, defined in Sec.\ref{sec:Static Obstacle Avoidance Contraints}, and $\bm{v}_{n_e+1} = \bm{v}_{1}$.
Due to the convexity of the model,
the distance between the moving obstacle and the hyperplane is converted into the minimum distance between  vertices and the hyperplane, 
thus simplifying Eq.(\ref{eq:disOpe}):
\begin{flalign}
	{\rm sd}(\bm{0},\mathbb{O}_{u}-\mathcal{P}_e) &= \frac{1}{\left \| \bm{{\rm H}}^e \right \|_2}(-{\rm h}^e + \min_{\bm o} {\left(\bm{{\rm H}}^e\right)}^{\rm T}\bm v^u_o ),
	\nonumber
	\\
	\bm v^u_o &= \bm{w}_u+\bm{{\rm R}}_u\bm l^u_o, \ \	o\in \{1,...,n_u\} \label{eq:overtics},
\end{flalign} 
where $\bm v^u_o$ is any vertex of the moving obstacle $\mathbb{O}_u$. $\bm{w}_u$ and $\bm{{\rm R}}_u$ are the origin and rotation matrix of the obstacle body coordinate system, respectively.  $l^u_o$ is the translation vector determined in advance.
Before further derivation, we define an auxiliary expression $\mathcal{H}(\widetilde{\bm{{\rm R}}},\Delta\bm{l}):(\mathbb{R}^{2\times2},\mathbb{R}^2)\rightarrow \mathbb{R}^2$:
\begin{flalign}
	\mathcal{H}(\widetilde{\bm{{\rm R}}},\Delta\bm{l}) = \frac{\bm{{\rm B}}\widetilde{\bm{{\rm R}}}\Delta\bm{l}}{\|\Delta\bm{l}\|_2} \label{eq:H}.
\end{flalign} 
By combining Eq.(\ref{eq:Evertics})(\ref{eq:He}-\ref{eq:overtics}), the signed distance can be analytically calculated:
\begin{flalign}
	{\rm sd}(\bm{0}&,\mathbb{O}_{u}-\mathcal{P}_e) = \min_{o} \mathcal{H}({{\bm{{\rm R}}},\Delta\bm{l}_e)}^{\rm T} (\bm{v}_o^u-\bm{v}_e), \label{eq:sd1}
\end{flalign}where $\Delta \bm{{l}}_e = \bm{l}_{e+1}-\bm{l}_e$, and the
physical meaning of $\mathcal{H}({{\bm{{\rm R}}},\Delta\bm{l}_e)}$ is the normal vector outwards of the plane $\mathcal{P}_e$. 
Similarly, the  signed distance between $\mathcal{P}_{o}^u$ and $\mathbb{E}$ can also be obtained:
\begin{flalign}
	{\rm sd}(\bm{0},\mathcal{P}_{o}^u-\mathbb{E}) = \min_{e}   \mathcal{H}(\bm{{\rm R}}_u,\Delta \bm{l}_o^u)^{\rm T}   \left(\bm{v}_e-\bm{v}_o^u\right) \label{eq:sd2}.
\end{flalign}
Finally, we substitute Eq.(\ref{eq:sd1})(\ref{eq:sd2}) into Eq.(\ref{eq:lowerbound}) to obtain the analytical expression of ${\rm lb_{sd}}$:
\begin{small}
\begin{flalign}
	&{\rm lb_{sd} }\left(\mathbb{E}(\bm{\sigma},\dot{\bm{\sigma}}), \mathbb{O}_{u}(\hat t)\right) =\nonumber\\
	& \max \{\max_{e}\min_{o} \mathcal{H}({{\bm{{\rm R}}},\Delta\bm{l}_e)}^{\rm T} (\bm{v}_o^u(\hat t)-\bm{v}_e)  ,\nonumber\\
	&\max_{o}\min_{e}  \mathcal{H}(\bm{{\rm R}}_u(\hat t),\Delta \bm{l}_o^u)^{\rm T}\left(\bm{v}_e-\bm{v}_o^u(\hat t)\right)\} \nonumber ,\\
	& \quad \quad e \in \{1,...,n_e\}, o\in\{1,...,n_u\} \label{eq:analowerbound}.
\end{flalign}\end{small}To smooth the maximum and minimum operations, a widely adopted log-sum-exp function is applied, defined as follows to approximate the vector-max(min) function:
\begin{flalign}
	{\rm lse}_{\alpha}(\bm{\gamma}) =  \alpha^{-1} \log \left (\sum_{\omega=1}^{\Omega} \exp{ (\alpha r_\omega )}\right),
\end{flalign}
where $r_{\omega}$ is a element of the vector $\bm{\gamma}=\{r_1,...,r_\omega,...,r_\Omega\}^{\rm T} \in \mathbb{R}^{\Omega>0}$.
if $\alpha > 0$ then it approximately gets the maximum value in $\bm{\gamma}$, or $\alpha < 0$ selects the minimum term. We smooth the discrete function by the log-sum-exp function with the advantage that the gradient of the log-sum-exp function is exactly the softmax(min) function. 
Moreover, the approximation error of the log-sum-exp is lower bounded by:
\begin{flalign}
	{\rm lse}_{\alpha>0}(\bm \gamma) \geq \max\{\bm \gamma\} \geq {\rm lse}_{\alpha>0}(\bm \gamma) - \frac{\log(\Omega)}{\alpha}.
\end{flalign}
Hence, we can formulate the distance function:
\begin{small}
\begin{flalign}
	U(\mathbb{E}(\bm{\sigma},\dot{\bm{\sigma}}),& \mathbb{O}_{u}(\hat t))=  {\rm lse}_{\alpha>0}(\bm{{\rm d}}) - \frac{\log(n_e+n_u)}{\alpha}, \nonumber\\
	\bm{{\rm d}} &= \left({\bm{d}_U}^{\rm T},{\bm{d}_E}^{\rm T}\right)^{\rm T} \in \mathbb{R}^{n_e+n_u}, \nonumber\\
	{\bm{d}_U} &= \left( d_U^1,...,{d}_U^e,...{d}_U^{n_e}\right)^{\rm T}\in \mathbb{R}^{n_e},\nonumber\\
	{\bm{d}_E} &= \left( d_E^1,...,{d}_E^o,...{d}_E^{n_u}\right)^{\rm T}\in \mathbb{R}^{n_u},\label{eq:Ueo}
\end{flalign}
\end{small}where  $d_U^e = {\rm sd}(\bm{0},\mathbb{O}_{u}(\hat t)-\mathcal{P}_e(\bm{\sigma},\dot{\bm{\sigma}}))$ and ${d}_E^o = {\rm sd}(\bm{0},\mathcal{P}_{o}^u(\hat t)-\mathbb{E}(\bm{\sigma},\dot{\bm{\sigma}}))$ are defined to simplify the formulation. 
Similarly, the minimum operation in $d_U^e$ and   ${d}_E^o$ can by approximated by the log-sum-exp function with $\alpha < 0 $. 
Combined with Eq.(\ref{eq:sd1})(\ref{eq:sd2}) We transform the distance into the flat-output space as:
\begin{small}
\begin{flalign}
	d_U^e &= {\rm lse}_{\alpha<0}\left(\left(d_{U_1}^e,...,d_{U_o}^e,...,d_{U_{n_u}}^e\right)^{\rm T} \right)+\widetilde d_U^e  , \nonumber \\
	d_E^o &= {\rm lse}_{\alpha<0} \left(\left(d^o_{E_1},...,d^o_{E_e},...,d^o_{E_{n_e}}\right)^{\rm T}\right)+\widetilde d_E^o ,\nonumber\\
	d_{U_o}^e &=   {\mathcal{H}({{\bm{{\rm R}}},\Delta\bm{l}_e)}}^{\rm T} \bm{{\rm R}}_u(\hat t)\bm{l}^u_o,\nonumber \\
	d_{E_e}^o &=   \mathcal{H}(\bm{{\rm R}}_u(\hat t),\Delta \bm{l}_o^u)^{\rm T}\bm{{\rm R}}\bm{l}_e\label{eq:deU}\nonumber,\\
	\widetilde d_U^e &= \mathcal{H}({{\bm{{\rm R}}},\Delta\bm{l}_e)}^{\rm T} (\bm{w}_u(\hat t)-\bm{v}_e),\nonumber \\	\widetilde d_E^o &= \mathcal{H}(\bm{{\rm R}}_u(\hat t),\Delta \bm{l}_o^u)^{\rm T}(\bm{\sigma} - \bm{v}_o^u(\hat t)). 
\end{flalign}
\end{small}By substituting Eq.(\ref{eq:deU}) into Eq.(\ref{eq:Ueo}) and then into Eq.(\ref{eq:dviolation}), $\mathcal{G}_{ \Theta_u}(\bm{\sigma},\dot{\bm{\sigma}}, \hat{t})$
is transformed into a continuously differentiable function that can be analytically expressed by  flat outputs.
Based on the chain rule, we  calculate the gradients:
\begin{small}
\begin{flalign}
	\frac{\partial \mathcal{G}_{ \Theta_u}}{\partial \bm{{\rm \sigma}}}& = 
	-\sum_{e=1}^{n_e}{\rm{lse}}_{\alpha>0}'(d^e_U)  \frac{\partial \widetilde d_U^e}{\partial \bm{\sigma}}
	-\sum_{o=1}^{n_u}{\rm{lse}}_{\alpha>0}'(d^o_E) \frac{\partial \widetilde d_E^o}{\partial \bm{\sigma}} ,\nonumber \\
	\frac{\partial \widetilde d_U^e }{\partial \bm{{\rm \sigma}}} & = -\mathcal{H}({{\bm{{\rm R}}},\Delta\bm{l}_e)} , \ \ 
	\frac{\partial \widetilde d_E^o}{\partial \bm{\sigma}} = \mathcal{H}(\bm{{\rm R}}_u(\hat t),\Delta \bm{l}_o^u),
\end{flalign}
\end{small}where ${\rm{lse}}':\mathbb{R}\rightarrow\mathbb{R}$ is the gradient of the log-sum-exp function.
In a similar way, we  derive the gradients w.r.t  $\hat t$ and $\dot{\bm{{\sigma}}}$:
\begin{small}
	\begin{flalign}
		\frac{\partial \mathcal{G}_{ \Theta_u}}{\partial \hat t} =&-\sum_{e=1}^{n_e}{\rm{{lse}}}'_{\alpha>0}(d^e_U)\left( \sum_{o=1}^{n_u} {\rm{{lse}}}'_{\alpha<0}(d^e_{U_o}) \frac{\partial d^e_{U_o}}{\partial \hat t} +\frac{\partial \widetilde d^e_{U}}{\partial \hat t}\right) \nonumber \\
		&-\sum_{o=1}^{n_u}{\rm{{lse}}}'_{\alpha>0}(d^o_E) \left(\sum_{e=1}^{n_e} {\rm{{lse}}}'_{\alpha<0}(d^o_{E_e}) \frac{\partial d^o_{E_e}}{\partial \hat t} \label{eq:ghat1}
		+\frac{\partial \widetilde d^o_{E}}{\partial \hat t}  \right)\nonumber,\\
		\frac{\partial \mathcal{G}_{ \Theta_u}}{\partial \dot{\bm{\sigma}}} =&-\sum_{e=1}^{n_e}{\rm{{lse}}}'_{\alpha>0}(d^e_U)\left( \sum_{o=1}^{n_u} {\rm{{lse}}}'_{\alpha<0}(d^e_{U_o}) \frac{\partial d^e_{U_o}}{\partial \dot{\bm{\sigma}}} +\frac{\partial \widetilde d^e_{U}}{\partial \dot{\bm{\sigma}}}\right) \nonumber \\
		&-\sum_{o=1}^{n_u}{\rm{{lse}}}'_{\alpha>0}(d^o_E)\sum_{e=1}^{n_e} {\rm{{lse}}}'_{\alpha<0}(d^o_{E_e}) \frac{\partial d^o_{E_e}}{\partial \dot{\bm{\sigma}}}.
	\end{flalign}
\end{small}From Eq.(\ref{eq:ghat1}), we can obtain the gradients of the dynamic safety constraint once the gradients of  auxiliary distances  are determined.
Next, we  derive the gradients w.r.t $\dot{\bm{\sigma}}$:
\begin{small}
\begin{flalign}
	\frac{\partial d_{U_o}^{e}}{\partial \dot{{\bm{\sigma}}}} &= \frac{\mathcal{F}(\Delta {\bm{l}}_e)  {\bm{{\rm B}}}^{\rm T}{\bm{{\rm R}}}_u(\hat t){\bm{l}}_o^u}{\|\Delta {\bm{l}}_e\|_2},\nonumber\\
	\frac{\partial \widetilde d^e_U}{\partial \dot{{\bm{\sigma}}}} &=\frac{\mathcal{F}(\Delta {\bm{l}_e)}{\bm{{\rm B}}}^{\rm T}({\bm w}_u(\hat t)- {\bm{v}}_e)+
		\mathcal{F}(\bm{l}_e){\bm{{\rm B}}}^{\rm T} \bm{{\rm R}} \Delta \bm{l}_e}{\|\Delta {\bm{l}}_e\|_2}, \nonumber \\
	\frac{\partial  d^o_{E_e}}{\partial \dot{{\bm{\sigma}}}} &=\mathcal{F}({\bm{l}_e})	\mathcal{H}(\bm{{\rm R}}_u(\hat t),\Delta \bm{l}_o^u).
\end{flalign}
\end{small}Then, the gradients w.r.t abstract timestamp $\hat t$ are also derived as follows:
\begin{flalign}
	\frac{\partial d_{U_o}^{e}}{\partial \hat t} &= \left(\dot{\bm{{\rm R}}}_u(\hat t) \bm{l}^u_o\right)^{\rm T} \mathcal{H}({{\bm{{\rm R}}},\Delta\bm{l}_e)},\nonumber\\
	\frac{\partial \widetilde d^e_U}{\partial \hat t} &=       \left({\dot{\bm{w}}_u(\hat t)}\right)^{\rm T} \mathcal{H}({{\bm{{\rm R}}},\Delta\bm{l}_e)}, \nonumber\\
	\frac{\partial  d^o_{E_e}}{\partial \hat t} & = \left(\bm{{\rm R}}\bm{l}_e\right)^{\rm T} \mathcal{H}(\dot{\bm{{\rm R}}}_u(\hat t),\Delta \bm{l}_o^u) \nonumber, \\
	\frac{\partial \widetilde d^o_E}{\partial \hat t}&= \mathcal{H}(\dot{\bm{{\rm R}}}_u(\hat t),\Delta \bm{l}_o^u)^{\rm T}
	(\bm{\sigma} - \bm{w}_u(\hat t)-\bm{{\rm R}}_u(\hat t)\bm{l}^u_o) \nonumber \\
	&-\mathcal{H}(\bm{{\rm R}}_u(\hat t),\Delta \bm{l}_o^u)^{\rm T} (\dot{\bm{w}}_u(\hat t)+\dot{\bm{{\rm R}}}_u(\hat t)\bm{l}^u_o),
\end{flalign}where $\dot{\bm{{\rm R}}}_u(\hat t)\in\mathbb{R}^{2\times 2}$ and $\dot{\bm{w}}_u(\hat t) \in \mathbb{R}^2$ are the gradients w.r.t $\hat t$.
In practice,  trajectories of other obstacles are fitted by piece-wise polynomials. Therefore, the position $\bm{w}_u(\hat t)$, the rotation matrix ${\bm{{\rm R}}}_u(\hat t)$ and their gradients can also be calculated analytically. Besides, we set $\alpha$ as $100$ to approximate the maximum function and $-100$ to smooth the minimum function.
\section{Reformulation of  Trajectory Optimization}
\label{sec:Reformulation of  Trajectory Optimization}
In this section, we analyze the characteristics of the constraints Eq.(\ref{eq:boundary}-\ref{eq:temporal})(\ref{eq:pointconstraint}) in trajectory planning and use targeted approaches to eliminate them, respectively.
Then, the original optimization problem is  reformulated  into  an unconstrained program that can be further solved efficiently.	
\subsection{Feasibility Constraints}	
We adopt the discrete-time summation-type penalty term  $S_{\Sigma}$ to relax the feasibility constraints Eq. (\ref{eq:pointconstraint}):
\begin{small}
\begin{equation}
	\begin{aligned}
		S_{\Sigma}({ \bm{{\rm c,T}}}) = \sum_{d\in \mathcal{D}}w_d\sum_{i=1}^{n}&\sum_{j=1}^{M_i}\sum_{k=0}^{\lambda}P_{d,i,j,k}(\bm{{c}}_{i,j},\bm{\mathrm{T}}),\\
		P_{d,i,j,k}(\bm{{c}}_{i,j},\bm{\mathrm{T}}) & = \frac{\delta T_i}{\lambda}  \bar{\omega}_k  \mathrm{L}_1(\mathcal{G}_{d,i,j,k}),\\
	\end{aligned}
	\label{eq:smapledpenalty}
\end{equation}
\end{small}where $w_d$ is the penalty weight corresponding to different kinds of constraints. $ [ \bar{\omega}_0, \bar{\omega}_1, ..., \bar{\omega}_{\lambda-1} ,\bar{\omega}_{\lambda} ] = [1/2,1, ... ,1,1/2]$ are the quadrature coefficients from the
trapezoidal rule\cite{alma9977072861303681} and $P_{d,i,j,k}$ is the violation penalty for a constraint point.
Moreover, we define a first-order relaxation function $\mathrm{L}_1(\cdot)$ to guarantee the continuous differentiability and non-negativity of  penalty terms:
\begin{flalign}
	\mathrm{L}_1(x) &=\left\{
	\begin{aligned}
		& 0					& x \leq 0 ,\\
		& -\frac{1}{2a_0^3}x^4+\frac{1}{a_0^2}x^3			& 0<x \leq a_0\\
		&x-\frac{a_0}{2}                & a_0<x.
	\end{aligned}
	\right.
\end{flalign}
Here $a_0=10^{-4}$ is the demarcation point. 
Such discrete penalty formulation ensures that continuous-time constraints Eq.(\ref{eq:user}) are satisfied within an acceptable tolerance. 
Then, the trajectory optimization for vehicles is reformatted as  follows:
\begin{subequations}
	\begin{align}
		\min_{ \bm{{\rm c,T}}}   \mathscr{J}({ \bm{{\rm c,T}}}) &= J({ \bm{{\rm c,T}}}) + S_{\Sigma}({ \bm{{\rm c,T}}})   
	\end{align}
	\setlength\abovedisplayskip{0.0pt}
	\begin{align}
		{\rm s.t.} 
		&\bm{\sigma}^{[s-1]}_0(0) = \bar{\bm{\sigma}}_{0}, \ \bm{\sigma}^{[s-1]}_n(T_n) = \bar{\bm{\sigma}}_{f}, \label{eq:boundary1} \\
		&\bm{\sigma}^{[s-1]}_i(T_i)  = \bm{\sigma}^{[s-1]}_{i+1}(0) = \widetilde{\bm{\sigma}}_{i}, 1\leq i <n, \label{eq:gearshift1} \\
		&\bm{\sigma}^{[\widetilde d]}_{i,j}(\delta T_i) = \bm{\sigma}^{[\widetilde d]}_{i,j+1}(0), 1\leq i \leq n, 1\leq j < M_i\label{eq:continuity1},\\
		&T_i > 0, 1\leq i \leq n.
	\end{align}
\end{subequations}
Without loss of generality, the gradients of  the violation penalty at each constraint point on the trajectory are derived:
\begin{flalign}
	\frac{\partial P_{d,i,j,k}}{\partial {\bm{{ c }}}_{i,j} } &= 
	\frac{\partial P_{d,i,j,k}}{\partial \mathcal{G}_{d,i,j,k} }
	\frac{\partial \mathcal{G}_{d,i,j,k}}{\partial {\bm{{ c }}}_{i,j} },\nonumber\\
	\frac{\partial P_{d,i,j,k}}{\partial \bm{{\rm T}}} &=
	\left(\mathbf{0}_{i-1}^{\rm T},\frac{P_{d,i,j,k}}{T_i},	\mathbf{0}_{n-i}^{\rm T}\right)^{\rm T}
	+  \frac{\partial P_{d,i,j,k}}{\partial \mathcal{G}_{d,i,j,k} } 
	\frac{\partial \mathcal{G}_{d,i,j,k}}{\partial  {\bm{\mathrm{T}}}}, \nonumber
	\\ \frac{\partial  P_{d,i,j,k}}{\partial \mathcal{G}_{d,i,j,k} } &=\frac{\delta T_i}{\lambda}  \bar{\omega}_k  \mathrm{L}_1'(\mathcal{G}_{d,i,j,k}). \label{eq:gradPtoG}
\end{flalign}
Since the gradients of the constraints  $\mathcal{G}_{d,i,j,k}$ have been  calculated in Sect. \ref{sec:Static States Constraints} and Sect. \ref{sec:Dynamic Avoidance}, we can also analytically obtain the gradients of the violation penalty
$\partial P_{d,i,j,k}$ by the above propagation chain Eq.(\ref{eq:gradPtoG}). Then, the gradients of the newly 
defined objective function $\mathscr{J}({ \bm{{\rm c,T}}})$ can be easily obtained in a similar way.
\subsection{Equality Constraints}
\label{sec:Equality Constraints}
Before discussing the elimination of the equality constraints Eq.(\ref{eq:boundary1})-(\ref{eq:continuity1}), we first decompose and reformulate Eq.(\ref{eq:gearshift1}):
\begin{subequations}
	\setlength\abovedisplayskip{-5.0pt}
	\begin{align}
		\bm{\sigma}_i(T_i)  &= \bm{\sigma}_{i+1}(0) = \bm{p}^g_i, \label{eq:gearp}  \\
		\bm{\sigma}_i^{(1)}(T_i)  &= -\bm{\sigma}_{i+1}^{(1)}(0) = \bm{v}^g_i, \label{eq:gearv}  \\
		\bm{\sigma}_i^{(\bar d)}(T_i)  &= \bm{\sigma}_{i+1}^{(\bar d)}(0) = \mathbf{0}_{2}\label{eq:geara}, \\
		||\bm{v}^g_i||_2 &= \bar v,\\
		\forall i  \in \{1,2,3,...,&n-1\},  \forall \bar d  \in \{2,...,s-1\}.
	\end{align}
\end{subequations}
$p^g_i$ is the gear shifting position and $v^g_i$ is the final velocity before the shift.
It is worth noting that the velocity direction is reversed before and after the shift and its magnitude is set to a small non-zero value $\bar v$ to prevent singularities during optimization. For instance, $\bar v$ is set to $0.05$ in the actual implementation. 
Based on the optimality condition proved in \cite{wang2021geometrically}, the minimum control effort piece-wise polynomial coefficients $\bm{{ c}}_i$ are uniquely determined by the intermediate waypoints $\bm{{ q}}_i$, the time interval of each piece, and the head and tail states:
\begin{align}
	\bm{{\rm M}}_i(T_i) \bm{{ c}}_i =  \bm{{\rm b}}_i, 1\leq i\leq n, \label{eq:optimal}
\end{align}
where $\bm{{\rm M}}_i(T_i)\in \mathbb{R}^{2M_is\times 2M_is}$ is an invertible banded matrix whose specific form can be found in \cite{wang2021geometrically}. $\bm{{\rm b}}_{i\in \{1...n\}} \in \mathbb{R}^{2M_is\times 2}$ is defined as follows:
\begin{small}
	\begin{align}
		\bm{{\rm b}}_1= &\left( \bar{\bm{\sigma}}_{0},\bm{{ q}}_{1,1},\mathbf{0}_{2\times \widetilde{d}},...,\bm{{ q}}_{1,M_1-1},\mathbf{0}_{2\times \widetilde{d}},
		p_1^g, v_1^g, \mathbf{0}_{2 \times (s-2)}\right)^{\rm T},\nonumber\\
		\bm{{\rm b}}_n= &\left( p_{n-1}^g, v_{n-1}^g, \mathbf{0}_{2 \times (s-2)},\bm{{ q}}_{n,1},\mathbf{0}_{2\times \widetilde{d}},...,
		\bm{{ q}}_{n,M_n-1},\mathbf{0}_{2\times \widetilde{d}},\bar{\bm{\sigma}}_{f}
		\right)^{\rm T}.\nonumber\\
		\bm{{\rm b}}_i= &\left( p^g_{i-1},v^g_{i-1},\mathbf{0}_{2 \times (s-2)},
		\bm{{ q}}_{i,1},\mathbf{0}_{2\times \widetilde{d}},...,\right.\nonumber\\
		&\left. \bm{{ q}}_{i,M_i-1},\mathbf{0}_{2\times \widetilde{d}},
		p_i^g, v_i^g, \mathbf{0}_{2 \times (s-2)}\right)^{\rm T}, 1<i<n,
	\end{align}
\end{small}where the  degree of continuity $\widetilde{d}$  is set to $2s-1$ to satisify the optimality condition. Based on Eq.(\ref{eq:optimal}), we  use the waypoints  $\bm{{\rm q}}=\left(\bm{{ q}}_1,...,\bm{{ q}}_n\right) \in\mathbb{R}^{2\times\sum_{i=1}^{n}(M_i-1)}$, the time set $\bm{{\rm T}}$, the gear shifting position $\bm{{\rm p}}^g=\left(\bm{{p}}^g_1,...,\bm{{ p}}^g_{n-1}\right)\in\mathbb{R}^{2\times(n-1)}$ and the velocity $\bm{{\rm v}}^g=\left(\bm{{v}}^g_1,...,\bm{{ v}}^g_{n-1}\right)\in\mathbb{R}^{2\times(n-1)}$as  decision variables in the optimization problem without sacrificing optimality, then  the  constraints Eq.(\ref{eq:boundary1})(\ref{eq:continuity1})(\ref{eq:gearp}-\ref{eq:geara})  naturally satisfy:
\begin{subequations}
	\begin{align}
		\min_{ \bm{{\rm q}},\bm{{\rm T}},\bm{{\rm p}}^g,\bm{{\rm v}}^g}   \mathcal{J}\left(\bm{{\rm q}},\bm{{\rm T}},\bm{{\rm p}}^g,\bm{{\rm v}}^g\right) &= \mathscr{J}\left({ \bm{{\rm c}}\left(\bm{{\rm q}},\bm{{\rm T}},\bm{{\rm p}}^g,\bm{{\rm v}}^g\right),\bm{{\rm T}}  }\right)
	\end{align}
	\setlength\abovedisplayskip{-5.0pt}
	\begin{align}
		{\rm s.t.} 
		||\bm{v}^g_i||_2 &= \bar v,1\leq i \leq n,\label{eq:nonv}\\
		T_i &> 0, 1\leq i \leq n.\label{eq:post}
	\end{align}
\end{subequations}
As stated in the work\cite{wang2021geometrically}, the propagation of gradients from polynomial coefficients to time and waypoints can be computed with linear complexity, which is not presented repeatedly in this paper.
Here, we derive the gradients  w.r.t the gear shifting position and the velocity:
\begin{small}
	\begin{align}
		\frac{\partial\mathcal{J}}{\partial \bm{{p}^g_{i}}}&=\left(\bm{{\rm M}}_i^{\rm -T}\frac{\partial \mathscr{J}}{\bm{{ c}}_i}\right)^{\rm T}e_{(2M_i-1)s+1}+\left(\bm{{\rm M}}_{i+1}^{\rm -T}\frac{\partial \mathscr{J}}{\bm{{ c}}_{i+1}}\right)^{\rm T}e_{1},\\
		\frac{\partial\mathcal{J}}{\partial \bm{{v}^g_{i}}}&=\left(\bm{{\rm M}}_i^{\rm -T}\frac{\partial \mathscr{J}}{\bm{{ c}}_i}\right)^{\rm T}e_{(2M_i-1)s+2}+\left(\bm{{\rm M}}_{i+1}^{\rm -T}\frac{\partial \mathscr{J}}{\bm{{ c}}_{i+1}}\right)^{\rm T}e_{2},\\
		\forall i &\in \{1,2,3,...,n-1\}.\nonumber
	\end{align}
\end{small}$e_k$ is  a column vector of proper dimension, where the element of the $k$-th row is $1$ and all others are $0$.
Furthermore, to satisfy the constraint Eq.(\ref{eq:nonv}), we express $\bm{v}^g_i=\left(\bar v\cos\theta^g_i, \bar v\sin\theta^g_i\right)^{\rm T}$ in polar coordinates, where the physical meaning of $\theta^g_i$ is the direction of velocity before gear switching.
Besides, we define the angle set $\bm{\mathrm{\theta}}^g = \left(\theta_1^g,\theta^g_2,...,\theta^g_{n-1}\right)^{\rm T}\in\mathbb{R}^{n-1}$ for subsequent derivation and the cost function  is transformed to $\mathcal{K}\left(\bm{{\rm q}},\bm{{\rm T}},\bm{{\rm p}}^g,\bm{{\rm \theta}}^g\right)=
\mathcal{J}\left(\bm{{\rm q}},\bm{{\rm T}},\bm{{\rm p}}^g,\bm{{\rm v}}^g\left(\bm{{\rm \theta}}^g\right)\right)$.
The gradient of the cost function $\mathcal{K}$ w.r.t $\bm{\theta}^g$ can be obtained as follows:
\begin{align}
	\frac{\partial\mathcal{K}}{\partial \theta_i^g}&=\left(-\bar v\sin\theta_i^g,\bar v\cos\theta_i^g\right)
	\frac{\partial\mathcal{J}}{\partial \bm{{v}}_i^g}, 1\leq i < n.
\end{align}
We  eliminate all equation constraints in the original planning which is the basis for subsequent efficient optimization.
\subsection{Positiveness Condition}
To remove the strict positiveness condition Eq.(\ref{eq:post}), we define an unconstrained virtual time  $\bm{\tau}=[ \tau_1,...,\tau_i,...,\tau_n]^{\rm T} \in \mathbb{R}^{n}$ and employ a diffeomorphism map~\cite{leslie1967differential} from   $\tau_i \in \mathbb{R}$ to the real duration $ T_i\in \mathbb{R}^{+}$:
\begin{equation}
	\begin{aligned}
		T_i &=\left\{
		\begin{aligned}
			& \frac{1}{2}\tau_i^2+\tau_i+1					& \tau_i > 0 \\
			& \frac{2}{\tau_i^2-2\tau_i+2}  					& \tau_i \leq 0\\
		\end{aligned}
		\right.\\
		\forall i & \in \{1,2,3,...,n\}.
	\end{aligned}\label{eq:virtoreal}
\end{equation}
Then, we use  $\bm{\tau}$ to replace the real time set $\bm{\mathrm{T}}$ so that the strict positiveness constraints Eq.(\ref{eq:post})  are  naturally satisfied.
The corresponding reformulated optimization problem is as follows:
\begin{align}
	\min_{ \bm{{\rm q}},\bm{{\rm \tau}},\bm{{\rm p}}^g,\bm{{\rm \theta}}^g}   \mathcal{W}\left(\bm{{\rm q}},\bm{{\rm \tau}},\bm{{\rm p}}^g,\bm{\mathrm{\theta}}^g\right) &=\mathcal{K}\left(\bm{{\rm q}},
	\bm{{\rm T}}\left(  \bm{\tau}  \right)
	,\bm{{\rm p}}^g,\bm{{\rm \theta}}^g\right).\label{eq:opt}
\end{align}
The gradient can be propagated from $ T_i$ to $\tau_i$:
\begin{equation}
	\begin{aligned}
		\frac{\partial \mathcal{W}}{\partial \tau_i} &=\left\{
		\begin{aligned}
			& (\tau_i + 1)\frac{\partial \mathcal{K}}{\partial T_i}					& \tau_i > 0 \\
			& \frac{4(1-\tau_i)}{{(\tau_i^2-2\tau_i+2)}^2}\frac{\partial \mathcal{K}}{\partial T_i} 					& \tau_i \leq 0\\
		\end{aligned}
		\right.\\
		\forall i & \in \{1,2,3,...,n\}.
	\end{aligned}
\end{equation}
The gradients w.r.t $\bm{\tau}$ are efficiently  calculated once we obtain the gradients  w.r.t $\bm{\mathrm{T}}$. 

In summary, we have removed all constraints in the optimization and analytically derived the  gradients. 
Finally, we robustly  solve the  transformed unconstrained optimization Eq.(\ref{eq:opt}) via the quasi-Newton method~\cite{liu1989limited}.

\begin{figure}[t]  
	\vspace{-0.0cm}  
	\centering
	{\includegraphics[width=1.0\columnwidth]{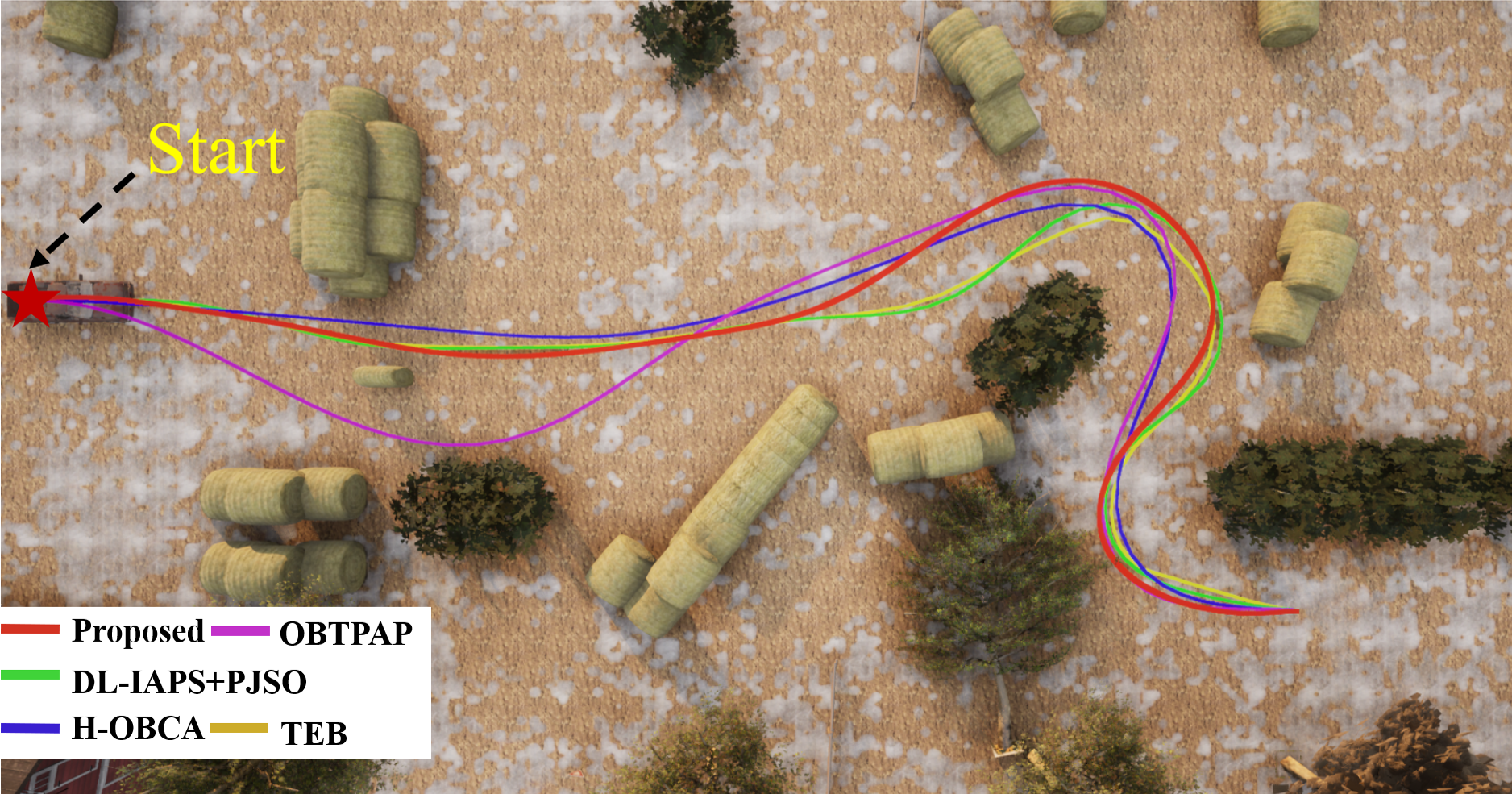}}
	\caption{ The trajectory visualization.
	}\label{fig:statictraj}
\end{figure}
\begin{figure}[t]  
	\vspace{-0.0cm}  
	\centering
	{\includegraphics[width=1.0\columnwidth]{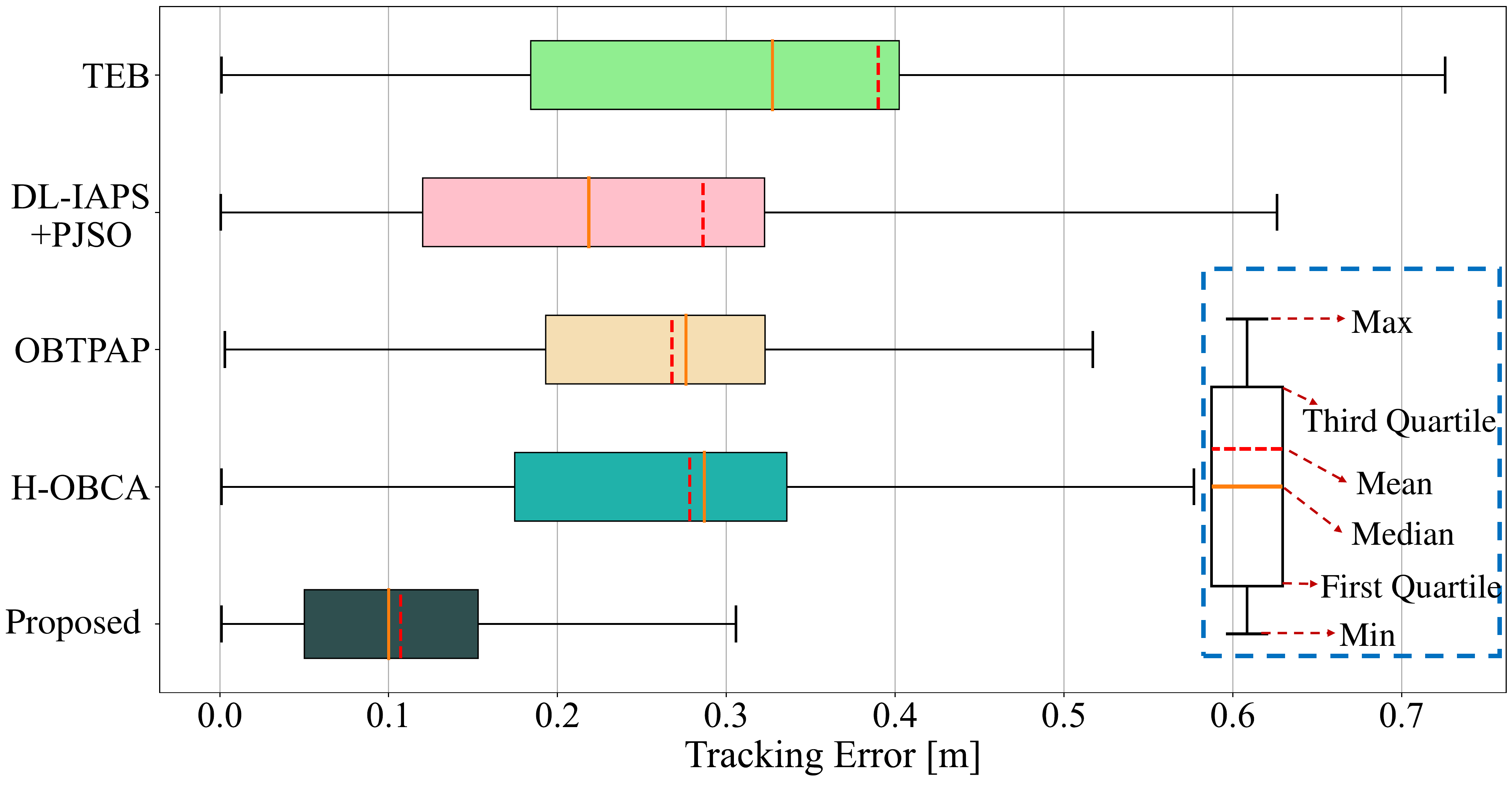}}
	\caption{Comparison of the tracking error profiles for the planned trajectories shown in Fig. \ref{fig:statictraj}.}
	\label{fig:trackprofile}
\end{figure}

\section{Evaluations}
\label{sec:Benchmarks}
The proposed method is evaluated both in simulation and in the real world.
Benchmarks show that our planner outperforms  state-of-the-art methods  in terms of time efficiency and trajectory quality.
\subsection{Simulations}
All  simulation experiments are conducted on a desktop computer running Ubuntu 18.04 with an Intel Core i7-10700 CPU and a GeForce RTX 2060 GPU. 
\begin{figure*}[t]  
	\vspace{-0.0cm}  
	\centering
	{\includegraphics[width=1.0\linewidth]{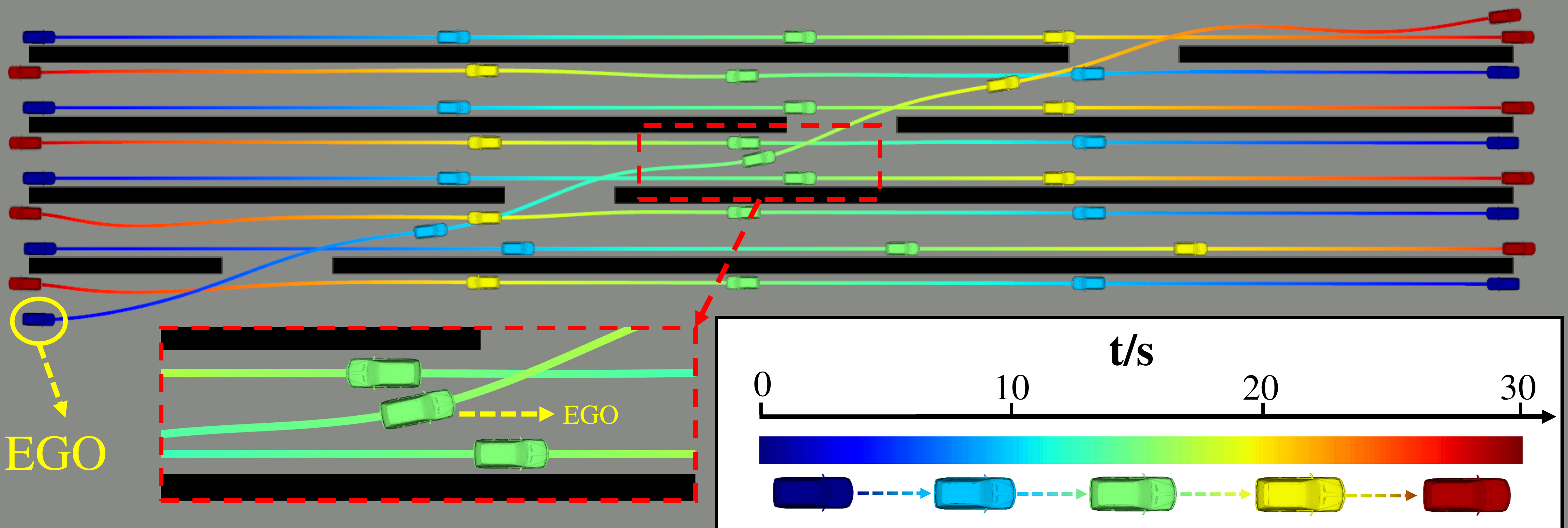}}
	\caption{Motion visualization in the dynamic environment, where the colors represent  timestamps.}
	\label{fig:dynSim}
\end{figure*}
\begin{figure}[t]  
	\vspace{-0.0cm}  
	\centering
	{\includegraphics[width=1.0\columnwidth]{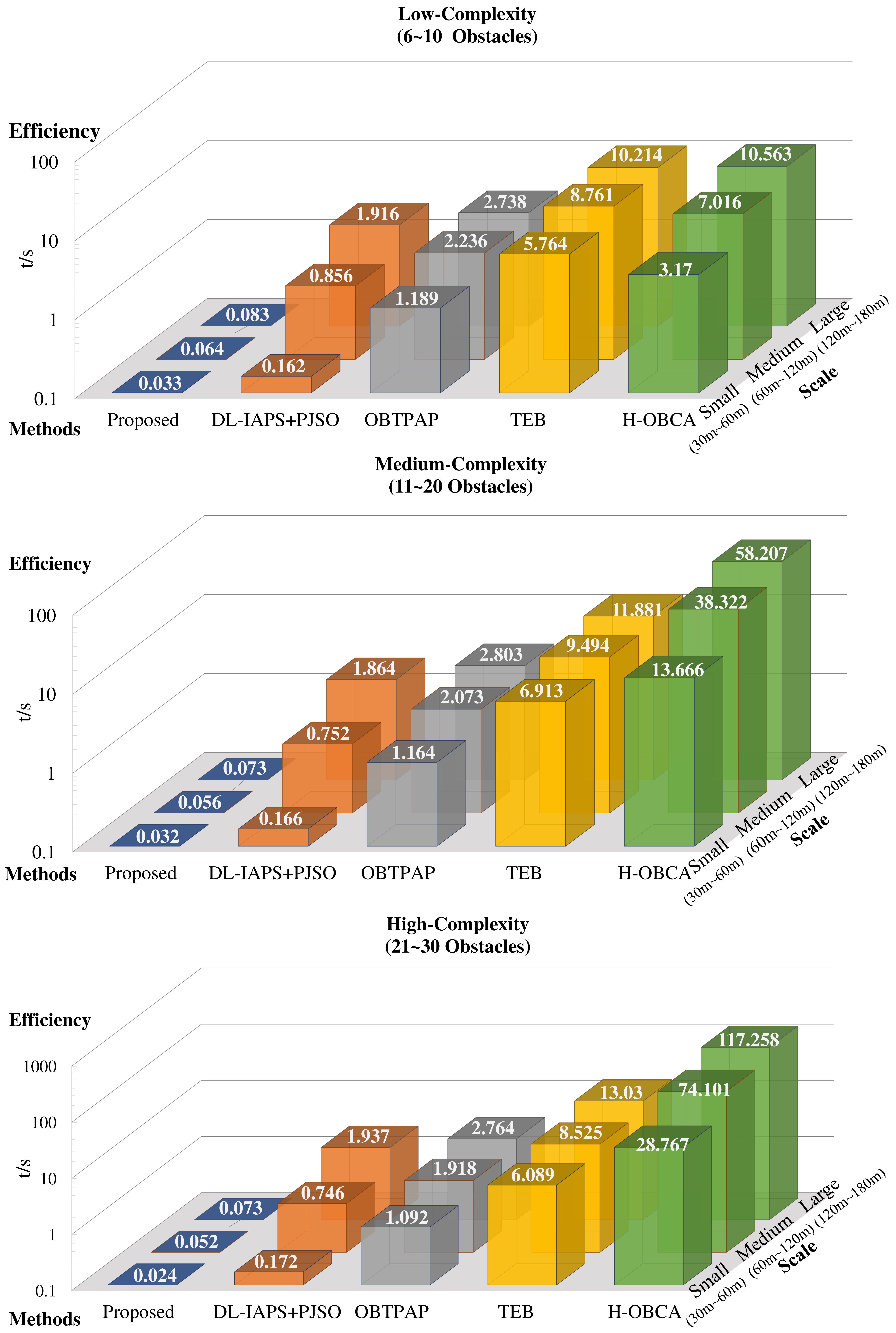}}
	\caption{Illustration of time efficiency comparisons.}
	\label{fig:timecomp}
\end{figure}
We perform  simulation experiments based on 
the open-source physical simulator CARLA\cite{dosovitskiy2017carla}.
The proposed method is compared with four impressive methods specifically designed for motion planning of car-like robots in static unstructured environments, including 
OBTPAP\cite{li2021optimization}, DL-IAPS+PJSO\cite{ZhouPJSO}, H-OBCA\cite{zhang2018autonomous} and Timed Elastic Bands (TEB) \cite{rosmann2017kinodynamic}.
All methods  use the bicycle model and are implemented in C++14 without  parallel acceleration.
OBTPAP \cite{li2021optimization} and H-OBCA\cite{zhang2018autonomous} are solved by primal-dual interior-point method IPOPT \cite{wachter2006implementation}.
DL-IAPS+PJSO\cite{ZhouPJSO} is implemented using OSQP\cite{stellato2020osqp}. The graph optimization solver $\mathrm{G}^2\mathrm{o}$\cite{5979949} is used for TEB\cite{rosmann2017kinodynamic}. Moreover, for fair comparisons, all planners use hybridA*\cite{dolgov2010path} algorithm as their front-end to provide rough initial guesses for the subsequent optimization. 
The planned trajectory visualization is shown in Fig. \ref{fig:statictraj}. 
Additionally, the proposed planner, OBTPAP\cite{li2021optimization}, DL-IAPS+PJSO\cite{ZhouPJSO} and TEB\cite{rosmann2017kinodynamic}
use known global point clouds in the environment to construct safety constraints.
H-OBCA\cite{zhang2018autonomous} uses known convex polygons manually extracted from obstacles to construct its optimization formulation.
All common parameters, including convergence conditions, dynamics constraints, and the ego vehicle dimensions, are set to the same  for fairness.
We use an MPC controller\cite{kong2015kinematic} that minimizes position and velocity errors to follow planned trajectories to  measure the executable performance. 
We draw the tracking error of planned trajectories in Fig. \ref{fig:trackprofile}, which visually demonstrates the superiority of the proposed method in terms of  physical feasibility.

We conduct extensively quantitative assessments in many cases with different  numbers of environmental obstacles and start-end distances (denote problem scale). Besides, the convergence tolerance of the optimization is set to $10^{-4}$.
Lots of comparison tests are performed in each case with random starting and ending states, and the average planning  time under each test case is visualized in Fig. \ref{fig:timecomp} to evaluate the real-time performance of planners. The histogram shows that our planner has an order-of-magnitude efficiency advantage over other algorithms, especially for large-scale trajectory generation. Furthermore, the results demonstrate the robustness against the problem scale and the number of environmental obstacles, thus allowing it to adapt to different scenarios. Moreover,  we record  physical dynamics statistics of planned trajectories including the mean of acceleration (abbreviated as M.A), jerk (M.JK) and tracking error (M.TE), as shown in Teb. \ref{tab:staticcomp}. The table show that our planner has the lowest control effort and the best  human comfort in all cases. Additionally, the parametric form of the trajectory  inherently guarantees the continuity of the state and its finite-dimensional  derivatives, which makes our trajectory easier to track than the above methods~\cite{li2021optimization,ZhouPJSO,zhang2018autonomous,rosmann2017kinodynamic} of discrete motion processes. 
It is worth mentioning that H-OBCA\cite{zhang2018autonomous} also generates high-quality trajectories with dynamics statistics close to ours due to the accurate modeling of the problem.
However, H-OBCA\cite{zhang2018autonomous} is inefficient and cannot be applied in real time, as shown in  Fig. \ref{fig:timecomp}.
The inherent problematic formulation of H-OBCA\cite{zhang2018autonomous} causes it to be not robust to dense environments and imposes an intolerable computational burden as the number of obstacles increases.

We also validate the proposed planner in a $210m \times 50m $ dynamic unstructured environment where the ego vehicle is required to avoid static obstacles and the traffic flow  consisting of eight other vehicles, as shown in  Fig. \ref{fig:dynSim}. Since perception is not the focus of this paper, the environment and the trajectories of other vehicles are known to the planner.  
The maximum velocity of all vehicles is set to $10m/s$.
The colors of vehicles and trajectories in Fig. \ref{fig:dynSim} represent motion timestamps, which indicate the absence of spatial-temporal intersections between the ego vehicle and other moving objects, demonstrating dynamic safety. 
Thanks to full-dimensional modeling of objects, the ego vehicle has the ability to fully utilize the safe space to get
 through narrow areas, as the close-up shows.                                                                                                                                                                      
The trajectory length of the ego vehicle is $260 m$, while the computation time is only $0.18s$, which demonstrates the efficiency of our planner in complex dynamic environments, especially for long-distance global trajectory generation.
\begin{table*}
	\small
	\centering
	\renewcommand\arraystretch{1.2}
	\caption{\label{tab:staticcomp} Comparison of Dynamic  Statistics in Different Cases }
	\begin{tabular}{c|clllllllll}
		\hline
		\multirow{3}{*}{Environments}                                             & Problem Scale          & \multicolumn{3}{c}{Small-Scale} ($30m\sim 60m$)                                                & \multicolumn{3}{c}{Medium-Scale} ($60m\sim 120m$)                                               & \multicolumn{3}{c}{Large-Scale ($120m\sim 180m$)}                                                \\ \cline{2-11} 
		& \multirow{2}{*}{Method} & \multicolumn{1}{c}{M.A}  & \multicolumn{1}{c}{M.JK} & \multicolumn{1}{c}{M.TE} & \multicolumn{1}{c}{M.A}  & \multicolumn{1}{c}{M.JK} & \multicolumn{1}{c}{M.TE} & \multicolumn{1}{c}{M.A}  & \multicolumn{1}{c}{M.JK} & \multicolumn{1}{c}{M.TE} \\
		&                         & \multicolumn{1}{c}{$(m/s^2)$} & \multicolumn{1}{c}{$(m/s^3)$} & \multicolumn{1}{c}{$(m)$} & \multicolumn{1}{c}{$(m/s^2)$} & \multicolumn{1}{c}{$(m/s^3)$} & \multicolumn{1}{c}{$(m)$} & \multicolumn{1}{c}{$(m/s^2)$} & \multicolumn{1}{c}{$(m/s^3)$} & \multicolumn{1}{c}{$(m)$} \\ \hline
		\multirow{5}{*}{\begin{tabular}[c]{@{}c@{}}Low-Complexity\\($6\sim 10$ Obstacles) \end{tabular}}    
		& Proposed                
		&      \textbf{ 6.96}                   &    \textbf{    16.61 }           &   \textbf{0.142}
		&         \textbf{6.76}                 &         \textbf{      18.81   }        &     \textbf{0.131}
		&              \textbf{ 6.03}           &     \textbf{ 15.30 }                       &         \textbf{0.146}
		\\
		& OBTPAP                 
		&      30.11                    &         101.99                 &       0.223                   
		&        51.32                  &             157.19             &            0.231             
		&        36.10                  &            105.56              &             0.250             
		\\
		& DL-IAPS+PJSO           
		&   10.02                       &        66.44                  &    0.208                      
		&               9.40           &               77.79           &    0.240                      
		&        10.14                  &    86.09                      &  0.252                        
		\\
		& H-OBCA                 
		&          7.42                & 20.67                         &  0.194                       
		&          8.23                &            24.11              &   0.217                       
		&           6.40               &            19.62              &     0.258                     
		\\
		& TEB                     
		&          16.27                &   380.52                       &    0.348                      
		&           15.74               &         427.35                 &   0.338                       
		&          14.96                &           363.07               &      0.321                    
		\\ \hline
		\multirow{5}{*}{\begin{tabular}[c]{@{}c@{}}Medium-Complexity\\ ($11\sim 20$ Obstacles)\end{tabular}} 
		& Proposed                
		&        \textbf{7.50}                  &        \textbf{20.69}&           \textbf{0.142}               
		&        \textbf{7.60}                  &        \textbf{18.66}        &      \textbf{0.135}                    
		&        7.17            &      \textbf{17.68}               &      \textbf{0.115}                    \\
		& OBTPAP                  
		&          42.27                &           145.89               &    0.239                      
		&           58.01               &           197.21               &     0.236                     
		&        49.94                  &           158.09               &       0.251                   \\
		& DL-IAPS+PJSO            
		&      11.53                    &       83.92                   &    0.223                     
		&       13.01                   &        119.75                  &      0.257                   
		&        14.26                  &         140.55                 &       0.254                   
		\\
		& H-OBCA                 
		&     7.85                     &      21.69                    &     0.216                     
		&     8.88                     &      26.25                    &    0.213                      
		&    \textbf{ 7.05   }                  &      20.57                    &       0.231                   
		\\
		& TEB                     
		&     18.72                     &     466.93                     &      0.348                    
		&    21.94                      &     614.42                     &         0.343                
		&        21.32                  &       707.91                   &       0.357                   
		\\ \hline
		\multirow{5}{*}{\begin{tabular}[c]{@{}c@{}}High-Complexity\\ ($21\sim 30$ Obstacles)\end{tabular}}   
		& Proposed                
		&          \textbf{7.53}             &         \textbf{17.95}             &     \textbf{0.127}                     
		&       \textbf{7.97}              &         \textbf{19.49}             &        \textbf{0.120  }                
		&       \textbf{6.97}            &     \textbf{17.56}              &         \textbf{0.126                 }
		\\
		& OBTPAP                  
		&         36.54                 &            128.29              &       0.230                   
		&            43.90              &               157.32           &        0.247                  
		&        57.03                  &                 193.75         &         0.251                 
		\\
		& DL-IAPS+PJSO            
		&             12.25             &       94.43                   &      0.210                    
		&            16.24              &       155.38                   &       0.259                   
		&                 13.92         &   134.74                       &       0.268                   
		\\
		& H-OBCA                  
		&             8.09                &      23.58                    &    0.195                      
		&                8.17           &  25.77                        &     0.230                     
		&           7.05                &    21.49                      &     0.248                     
		\\
		& TEB                     
		&       19.67                   &            493.70              &   0.352                       
		&                23.42          &     607.32                     &     0.416                     
		&               19.89           &           584.01               &       0.365                   
		\\ \hline
	\end{tabular}
\end{table*}
\subsection{Real-World Experiments}
\label{sec:Real-World Experiments}
\begin{figure}[t]  
	\vspace{-0.0cm}  
	\centering
	{\includegraphics[width=1.0\columnwidth]{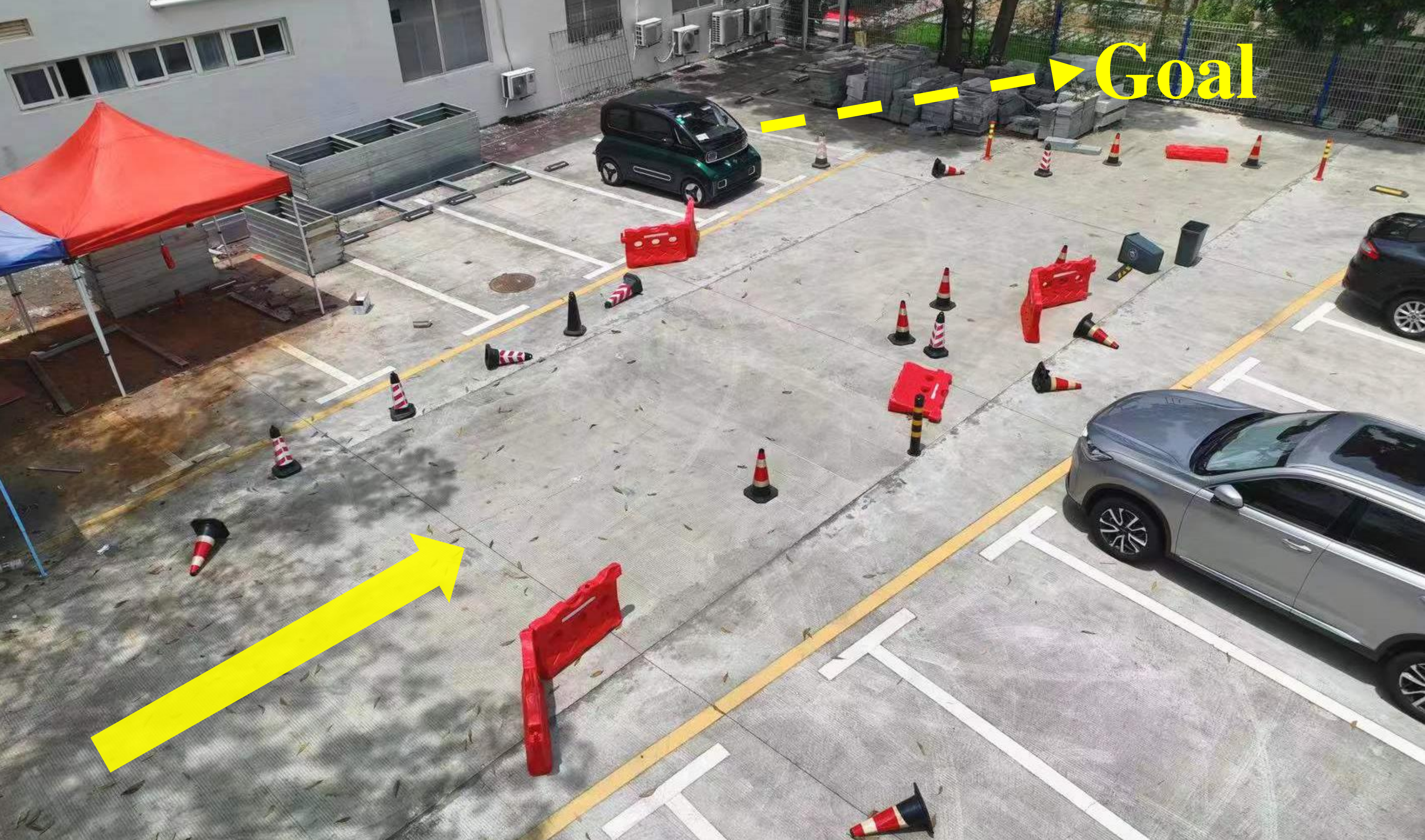}}
	\caption{The real-world experiment site.}
	\label{fig:realworld}
\end{figure}

\begin{figure}[t]  
	\vspace{-0.0cm}  
	\centering
	{\includegraphics[width=1.0\columnwidth]{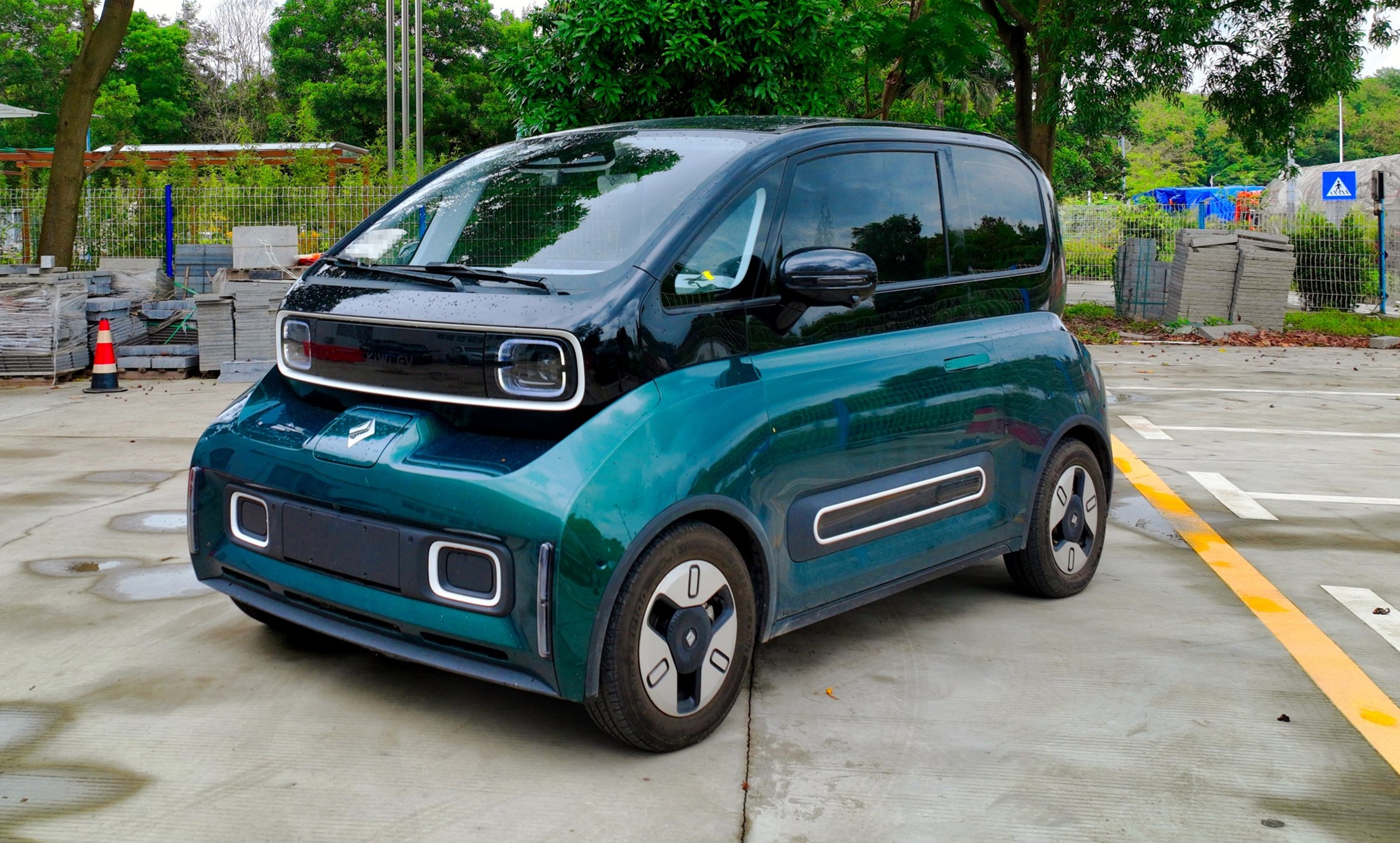}}
	\caption{The experimental platform.}
	\label{fig:wulin}
\end{figure}

In addition to simulations, we also conduct real-world experiments to verify the feasibility of our planner on a real physical platform.
The experimental site is a dense $40m \times 20m$ outdoor unstructured parking lot, as shown in Fig. \ref{fig:realworld}.
The moving robot is required to start from an initial state, go around obstacles, and eventually reverse into a parking space at a human-defined heading angle.
The length of the whole track is about 42 meters.
Real-world experiments are conducted on a SAIC-GM-Wuling Automobile Baojun E300\footnote{\url{https://www.sgmw.com.cn/E300.html/}} with dimensions of 
$2.9m \times1.7m \times1.6m $, a wheelbase of $2m$ and no GPS included, as shown in Fig. \ref{fig:wulin}.
A set of sensory elements including one  stereo camera, four  fisheye cameras, and 
twelve ultrasonic sensors  are deployed on the  platform for real-time localization, control, and data recording.
Additionally, neither high precision LIDAR nor  precise external positioning systems are used in our real-world experiments.
\begin{figure*}[t]  
	\vspace{-0.0cm}  
	\centering
	{\includegraphics[width=1.0\linewidth]{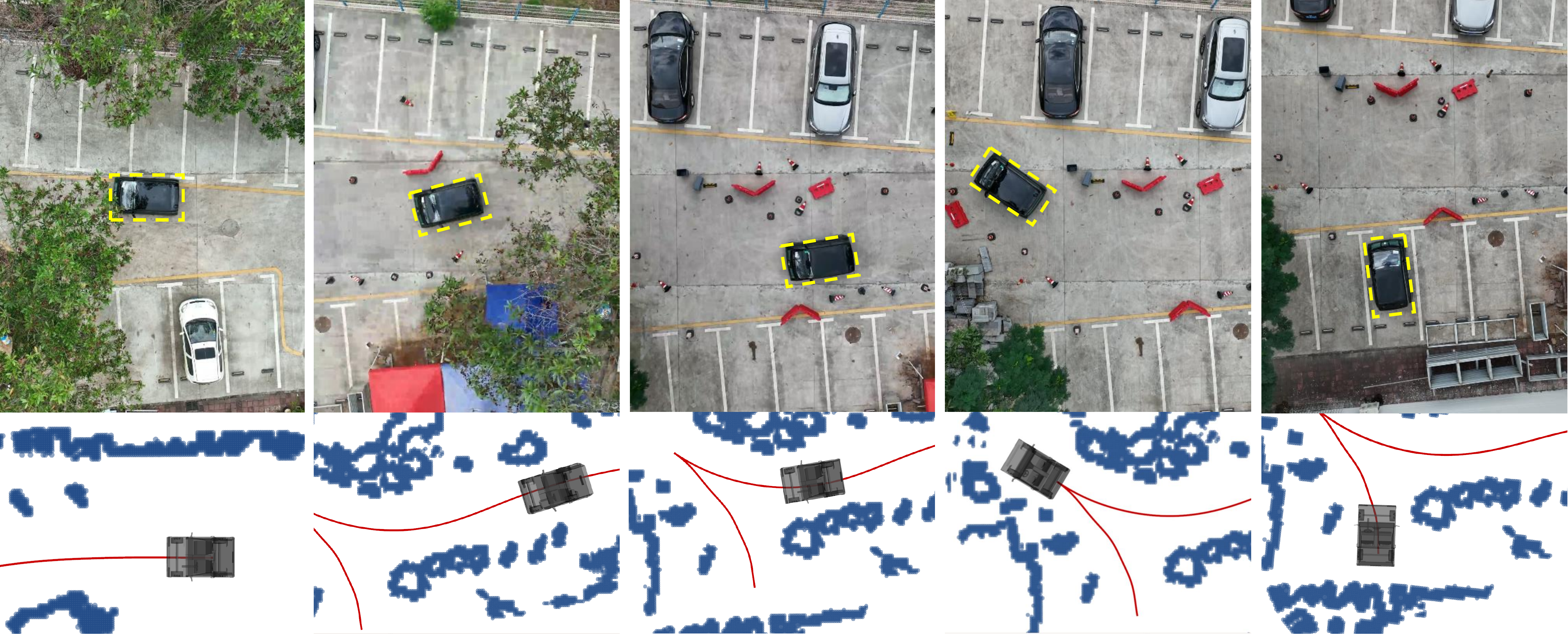}}
	\caption{The motion diagram of the ego vehicle in the real-world experiment, where the red curve is the execution  trajectory.}
	\label{fig:realfig2}
\end{figure*}
All modules are implemented in C++ and deployed on TI TDA4 Chip (64-bit Dual Arm Cortex-A72 @2000MHz) \footnote{\url{https://www.ti.com/product/TDA4VM}} configured with QNX Operating System.
During the trajectory planning, the vehicle shape was expanded by $0.25m$ to ensure  safety  during actual execution in the presence of unavoidable control and positioning errors.
The maximum forward and backward speeds are set to $2m/s$ and $0.5m/s$, respectively.
The time weight $w_T$ is set to $50$ to ensure the aggressiveness of the trajectory.
The motion of the ego vehicle is visualized in Fig. \ref{fig:realfig2}, which demonstrates that the ego vehicle can follow the planned trajectory to avoid obstacles and eventually reverse into a parking space with the user-given heading angle.
Furthermore,  dynamic evaluation metrics are quantified  in Tab. \ref{tab:realtab}.
As we can see, the ego vehicle maintains a relatively high speed throughout to reach the target state without exceeding the dynamic limits while keeping a low jerk to ensure the comfort of  passengers.
Finally, more  demonstrations can be found in the supplementary material.

\begin{table}[t]
	\small
	\centering
	\renewcommand\arraystretch{1.15}
	\caption{\label{tab:realtab} Dynamic Statistics in Real-World Experiments }
	\begin{tabular}{lclll}
		\toprule
		Statistics & Mean  &  Max & STD.\\
		\midrule
		Forward.Vel. ($m/s$) &1.53  &1.95 &0.52	\\
		\midrule
		Backward.Vel. ($m/s$) & 0.45 &0.50  &0.10\\
		\midrule
		Forward.Jerk. ($m/s^3$) & 0.25 &6.95  &0.50 \\
		\midrule
		Backward.Jerk ($m/s^3$)&0.18  & 7.07 &0.55 \\
		\bottomrule
	\end{tabular}
\end{table}
\section{Conclusion}
\label{sec:conclusion}
This paper proposes an efficient spatial-temporal trajectory planning for  car-like robots.
We use polynomials in flat space to parameterize trajectories such that fewer decision variables characterize a continuous trajectory with higher-order information. 
Moreover, the spatial-temporal joint optimization of the trajectory also guarantees better optimality. 
We decompose  ambient point clouds and free space to construct a safe driving corridor, thus modeling  geometric constraints used to ensure static obstacle avoidance.
To cope with dynamic environments, we constrain  the signed distances between the ego vehicle and moving objects to achieve  full-dimensional obstacle avoidance. Benchmark results with state-of-the-art methods demonstrate the superiority of our method in terms of time efficiency and trajectory quality. Real-world experiments are also conducted to validate the effectiveness of the proposed method on a real platform.

In the future, we will explore the application of the method to multiple robots and extend the principle to  other motion models. In addition, we will focus on motion planning for rugged terrain in the field, such as going up and down hills.


\bibliography{references}

\end{document}